\documentclass[11pt]{article}

\usepackage[final]{acl}

\usepackage{amsmath}

\usepackage{tikz}
\usetikzlibrary{positioning}

\usepackage{times}
\usepackage{latexsym}
\usepackage{booktabs}
\usepackage{listings}

\usepackage{bm}
\usepackage{amssymb}
\usepackage[T1]{fontenc}

\usepackage[utf8]{inputenc}

\usepackage{microtype}

\usepackage{inconsolata}

\usepackage{graphicx}
\usepackage{subcaption}

\usepackage{titlesec}

\titlespacing*{\section}
{0pt}{10pt plus 3pt minus 2pt}{6pt}

\titlespacing*{\subsection}
{0pt}{8pt plus 2pt minus 2pt}{4pt}

\titlespacing*{\subsubsection}
{0pt}{6pt plus 2pt minus 1pt}{3pt}
\usepackage{pgfplots}
\usepgfplotslibrary{statistics}
\pgfplotsset{compat=1.18}

\title{Graph Visualizations in Prompts for Knowledge Graph Question Answering }

\author{%
Luiz do Valle Miranda\textsuperscript{1},
Andreas L. Opdahl\textsuperscript{2},
Grzegorz J. Nalepa\textsuperscript{1}\\[1ex]
\textsuperscript{1}Jagiellonian University, Krak\'{o}w, Poland\\
\textsuperscript{2}University of Bergen, Bergen, Norway\\
\texttt{andreas.opdahl@uib.no}
}

\begin{document}
\pagenumbering{gobble}
\maketitle
\begin{abstract}
We study when and how multimodal input improves knowledge graph question answering (KGQA) in large vision--language models. Using a controlled evaluation over 1,500 WebQSP and CWQ questions and six VLMs within a fixed KGQA pipeline, we compare text-only prompting against text-plus-graph visualization at the per-question level and analyze performance changes as a function of reasoning complexity, model size, and image layout. Adding a graph image yields an average gain of 1.85 F1 points, driven by strong precision improvements on low-complexity questions (+4.65 F1). Mixed-effects and regression analyses show that these multimodal gains do not depend strongly on model scale, but are highly sensitive to graph layout: image-only settings sharply degrade performance, hierarchical Graphviz layouts outperform less organized renderings, and decomposing large subgraphs into separated paths yields the largest improvements. Taken together, the results indicate that visual prompting primarily aids KGQA by enhancing prediction precision in low-complexity contexts, provided that visual design respects cognitive and perceptual constraints. To optimize both accuracy and inference efficiency, we argue for adaptive visual prompting policies that selectively append diagrams to small, visually sparse subgraphs with well-structured layouts, avoiding unnecessary visual clutter and token overhead on complex queries. We conclude by relating these findings to a visual notation design theory, outlining how principled diagram encoding can guide future multimodal VLM prompting.
\end{abstract}

\section{Introduction}
Knowledge graphs (KGs) have become a widely used formalism for representing structured knowledge across domains such as the Semantic Web, digital libraries, and scientific information management \cite{hogan2021knowledge}. They organize information as entities and relations, supporting the explicit encoding of relational facts while preserving the connectivity and topology of the underlying data. At the same time, precisely because knowledge is encoded in a graph structure, it can be difficult for users to access, retrieve, and reason over KGs directly, especially without specialized query languages such as SPARQL.

Language models (LMs) have therefore been proposed as a natural-language interface to structured knowledge \cite{pan2023_KGandLLM}. A growing body of work has explored how LLMs can query, interpret, and reason on graph-based data, including link prediction \cite{ye-etal-2024-language} and node classification \cite{chen2024_node_class}. Some approaches focus specifically on KGs through text-to-SPARQL systems, where LMs convert a natural language question into a SPARQL query to be executed over a KG \cite{kosten2025_sparql} and retrieval-based KG Question Answering (KGQA), where the model uses natural language questions to retrieve triples of a KG, that are used as in-context learning for Q\&A \cite{baek2023_kaping,luo_reasoning_2024}.

In parallel, an emerging line of research investigates how LMs and vision language models (VLMs) can benefit from visual representations of structured information, such as charts, diagrams and other schematic visualizations \cite{zhang2024_surveyvlm}. VLMs have shown promising capabilities in chart and diagram understanding tasks, indicating that visual encodings can support fine-grained quantitative and relational reasoning \cite{zou-etal-2024-vgbench}. Graphs are particularly interesting in this context because they are not only a formal data structure, but can also be rendered visually as node–link diagrams, making relational patterns, connectivity, and local neighborhoods easier to perceive \cite{li2024visiongraph}.

Recent systems have begun to explicitly include graph visualizations in VLMs prompts, with the goal of improving graph-structured reasoning beyond what is possible with text alone. \cite{wei2024_GITA} presents GITA, a framework for fine-tuning LLMs to benefit from graph visual representations for graph reasoning tasks\footnote{Throughout this paper, ``reasoning'' refers to using available
textual, visual, and graph-structured evidence to derive an answer. The term
describes task-level performance rather than models' internal reasoning
processes or chain-of-thought strategies.}
Furthermore, \cite{deng_2025graphvis} introduce GraphVis, a method for using subgraphs and their image visualizations for fine-tuning VLMs towards improving their response in text-based Q\&A tasks. 

However, both approaches primarily study graph visualizations as part of a fine-tuning pipeline, where textual and visual graph representations are jointly used to adapt the model to graph-based reasoning tasks. Although fine-tuning can improve task-specific performance, it does not isolate whether the visual rendering itself provides useful information at inference time. In contrast, this paper explores this question through in-context learning, which allows us to evaluate the contribution of visual graph representations without updating model parameters or introducing fine-tuning.

To the best of our knowledge, the effect of providing a KG as a visual input to a VLM has not yet been systematically studied in isolation. In particular, prior work rarely compares a rendered KG directly against a textually equivalent list of triples while keeping both the model and the underlying knowledge constant. In this paper, therefore, we investigate the following research question:

\begin{quote}
\textbf{RQ:} Does including a KG visualization on inference time improve a VLM's performance in the KGQA task compared to providing the same graph only as a textual list of triples? 
\end{quote}

We evaluate a range of vision language models comparing performance under text-only and text-plus-graph (so-called multimodal) input. We analyze how visual representations of KGs affect Q\&A performance at different levels of complexity, model capability, and image style, providing an assessment of when and how graph visualizations contribute to improved KG understanding.

\vspace{-5pt}

\section{Methods}
\subsection{The KGQA task}
\begin{figure*}
    \centering
    \includegraphics[width=0.90\linewidth]{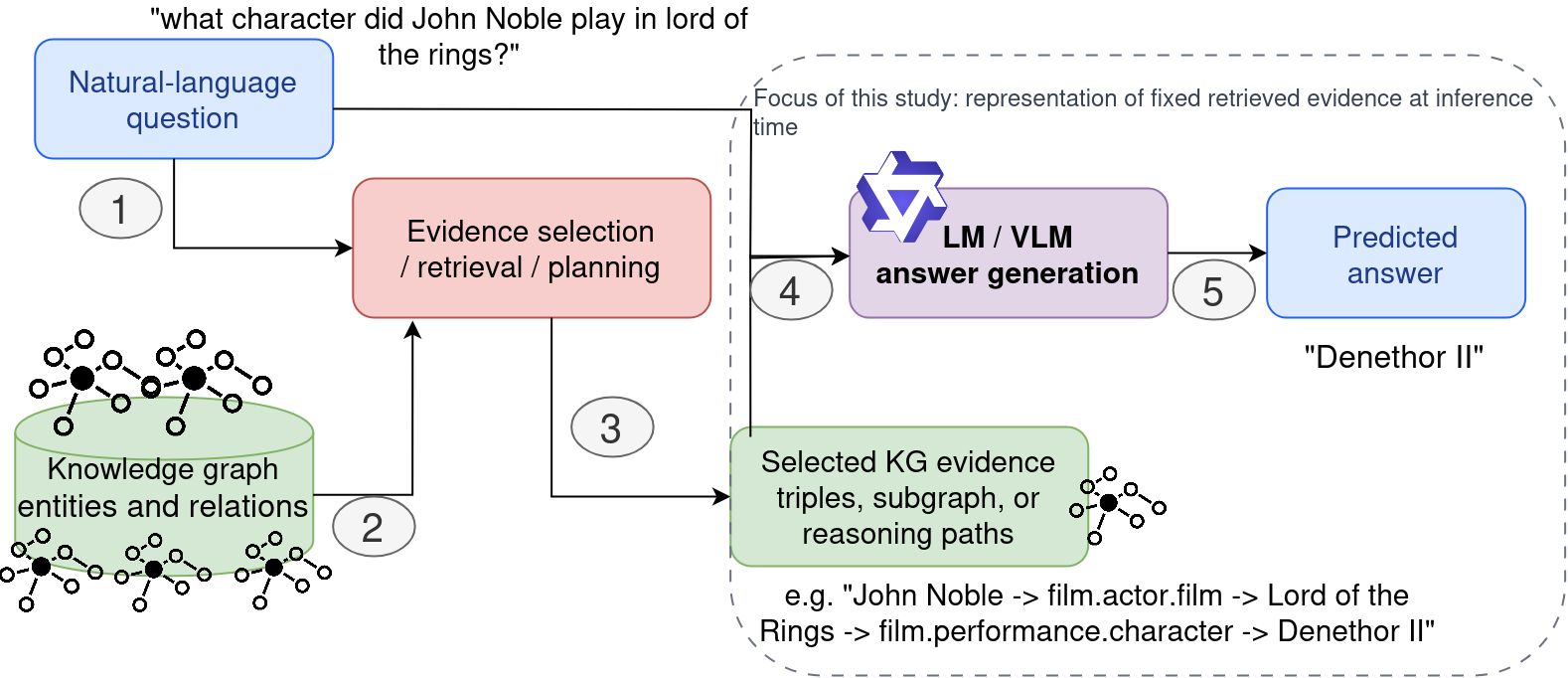}
    \caption{High-level abstraction of KGQA pipelines. Steps 1--3 retrieve relevant KG evidence;
Steps 4--5 use it to generate an answer. Our analysis is agnostic to 1--3. We focus on Step 4--5, comparing
textual and multimodal evidence representations at inference time.}
    \label{fig:kgqa_pipeline}
\end{figure*}

As the main experimental setting of this paper, we use the KGQA task, since it provides a natural setting for studying how language models use structured knowledge expressed as entities, relations, and triples. A foundational architecture for KGQA is KAPING \cite{baek2023_kaping}, a retrieval-based framework in which the input question is used to retrieve relevant facts from the KG, and these retrieved triples are then prefixed to the prompt of the language model to support answer generation. A development in KGQA task is Reasoning on Graphs (RoG) \cite{luo_reasoning_2024}, a framework in which an LM first generates relation paths as reasoning plans, then retrieves matching reasoning paths from the KG, and finally reasons over these paths to answer the question. A similar framework is Think-on-Graph \cite{sun2024_TOG}, a graph-exploration framework in which the LM collects evidence from a KG to construct reasoning paths during inference and then uses the explored graph evidence for answer generation.

Despite these differences, the three frameworks share a common pattern. A natural-language question is first used to identify relevant graph evidence, such as retrieved triples, subgraphs, or reasoning paths, and this structured evidence is then converted into textual input for the LM. The LM finally uses the question together with this graph-derived context to generate an answer. Fig. \ref{fig:kgqa_pipeline} presents a visualization of this pipeline.

In the following, we focus on the final stage of this pipeline, namely the interaction between graph-derived context and the LM during answer generation. Our central hypothesis is that augmenting the prompt with a visualization of the retrieved KG context can improve performance beyond using a textual representation of the same information alone. By isolating this stage, we can study the effect of the visual modality independently of the previous retrieval or planning components, making the hypothesis in principle applicable across different KGQA frameworks. For the concrete experimental setup, we adopt RoG as the base framework, since its modular planning--retrieval--reasoning design makes the answer-generation stage easier to isolate and analyze, while remaining simpler than more interactive approaches such as ToG.

\subsection{Dataset}
\begin{table}[t]
\centering
\footnotesize
\setlength{\tabcolsep}{3pt}
\caption{Evaluation dataset statistics.}
\label{tab:dataset-statistics}
\begin{tabular}{lrrr}
\toprule
 & WebQSP & CWQ & All \\
\midrule
Questions & 750 & 750 & 1,500 \\
Mean answers & 16.52 & 2.12 & 9.32 \\
Mean triples & 18.47 & 20.54 & 19.51 \\
Median triples & 9 & 14 & 12 \\
Low complexity ($\leq$7) & 354 & 282 & 636 \\
Medium complexity (8--25) & 159 & 184 & 343 \\
High complexity ($>$25) & 237 & 284 & 521 \\
\bottomrule
\end{tabular}
\end{table}

Following RoG, we evaluate VLMs on two benchmark KGQA datasets: WebQuestionSP (WebQSP) \cite{yih-etal-2016-value} and Complex WebQuestions (CWQ) \cite{talmor-berant-2018-web}. Both datasets contain natural-language questions paired with the appropriate Freebase nodes that successfully answer the question \cite{Bollacker2008_freebase}, making them standard benchmarks for KG-based question answering. They also allow controlled evaluation of multi-hop reasoning, with WebQSP containing relatively simpler questions and CWQ including more complex and challenging ones. 

Neither CWQ nor WebQSP make assumptions about the paths and strategies for KGQA. Thus, to construct our evaluation set, we use the reasoning paths made openly available by \cite{luo_reasoning_2024}\footnote{\url{https://github.com/RManLuo/reasoning-on-graphs}}. From this resource, we derive a balanced dataset of 1,500 question--answer pairs, with 750 instances from WebQSP and 750 from CWQ. The resulting dataset covers a range of graph contexts with varying numbers of triples, enabling evaluation at different levels of visual structural complexity.  
Complexity is operationalized based on the number of triples in each question, serving as a proxy for the length and structural depth of the required reasoning chain. The differences observed between complexity levels should be interpreted as reflecting the combined effects of evidence volume, reasoning requirements, and visual density.

We categorize instances with at most seven triples as \emph{low complexity}, those with 7--25 triples as \emph{medium complexity}, and those with more than 25 triples as \emph{high complexity}. This operationalization also correlates with the amount of visual content provided, as higher-complexity questions tend to be associated with larger or more information-dense input images.  
Table~\ref{tab:dataset-statistics} presents the statistics of the resulting dataset. Our dataset can be accessed in \url{https://zenodo.org/records/21620848} and our code in \url{https://github.com/luizdovalle2/VLM_KGQA}.

\subsection{Main experiment design}
To evaluate the main research question of this paper, we conducted experiments under two input conditions: one in which a VLM receives only a textual representation of the graph context, and one in which it receives both the textual representation and an image visualization of the same graph using Graphviz \cite{ellson2001_Graphviz}. An example of a graph visualization can be found in Fig \ref{fig:graphviz_1}.

We evaluate these conditions across a diverse set of models that vary in size, provider, and accessibility, including both open-weight and closed-weight systems, in order to examine whether the observed effect is consistent across model families rather than specific to a single architecture.
Among the evaluated models, we include two representatives of different sizes from three major model families: Qwen3.5-35B-A3B and Qwen3.5-122B-A10B; Mistral Small 3.1 24B  and Mistral Medium 3.1; and Gemini 3.1 Flash Lite together with Gemini 3 Flash. The used prompt template can be found in Appendix \ref{app:prompts}. We disabled the models' explicit reasoning mode to standardize answer generation across conditions.

Following \cite{luo_reasoning_2024}, we use F1, precision, and recall as evaluation metrics\footnote{We do not include  Hits@1  because it assumes a ranked list of candidate answers, while in our setting, the model produces an unordered set of answers. }. Because some questions have multiple valid answers, we treat both the predicted answers and the gold answers as sets. For each question, let $A_p$ denote the predicted answer set and $A_g$ the corresponding gold answer set. Precision and recall are computed as

\[
P = \frac{|A_p \cap A_g|}{|A_p|},
\qquad
R = \frac{|A_p \cap A_g|}{|A_g|},
\]

\noindent where the intersection represents the answers correctly predicted by the model.
We compute these metrics independently for each question and then macro-average the resulting per-question scores across the evaluation set.

We analyze performance across several dimensions of the task and input context. In particular, we examine the effect of model size, the number of triples in the graph context, and the complexity of the associated reasoning paths. These analyses help us better understand under which conditions visual graph representations are most beneficial.

\subsection{Construction of Graph-Derived Context}
\label{sec:graph_context_construction}

The graph-derived context used in our evaluation is constructed directly from the reasoning paths released by the RoG framework \citep{luo_reasoning_2024}. Each reasoning path consists of alternating entity and relation chains. We decompose these paths into distinct subject--predicate--object triples by extracting consecutive entity--relation--entity segments and removing duplicates.

This triple set constitutes the fixed evidence across all experimental conditions, ensuring that performance changes stem strictly from the visual representation rather than differing context scope. In the \emph{Text-Only} condition, triples are linearized sequentially into structured text strings. In the multimodal (\emph{Text + Image}) condition, the exact same context is additionally rendered as a directed visual graph using Graphviz's \texttt{dot} layout engine, where entities are displayed as rounded rectangular nodes and relations as labeled, directed edges in a left-to-right orientation. Complete pipeline elements, including reasoning paths, extracted triples, and corresponding image outputs, are detailed in Appendices~\ref{sec:case1_analysis}--\ref{sec:case3_analysis}.

\subsection{Ablation Design}
\label{sec:ablation_design}

To isolate the structural and perceptual factors driving multimodal performance changes, we conduct a series of controlled ablation studies. To eliminate confounding variance across distinct model architectures, all ablation experiments are evaluated on \texttt{Qwen3.5-35B-A3B}.

We examine four input presentation strategies:
\begin{enumerate}
    \item \textbf{Alternative Graph Layouts:} We compare the default hierarchical \texttt{dot} layout against two non-hierarchical spatial policies: an unstructured Graphviz layout and a force-directed radial layout using Networkx. This isolates whether downstream performance depends on graph topology visibility or simply the presence of images.
    \item \textbf{Reasoning Path Decomposition:} Rather than compiling all evidence into a single dense diagram, we partition subgraphs into an array of isolated single-chain path images. This tests whether reducing clutter improves joint constraint reasoning in complex subgraphs.
    \item \textbf{Image-Only Prompting:} We remove the linearized textual triples, providing only the graph visualization alongside the question to test whether visual diagrams can replace textual evidence.
\end{enumerate}

More information on rendering parameters and algorithm definitions for each layout variant are detailed in Appendix~\ref{sec:methodology_construction}. Qualitative visual comparisons of these presentation policies are illustrated in Figures~\ref{fig:case1_graphs}, \ref{fig:case2_graphs}, and \ref{fig:case3_graphs}.

\section{Results}
\subsection{Text-only vs multimodal comparison}

\begin{table}[t]
\centering
\small
\setlength{\tabcolsep}{4pt}
\renewcommand{\arraystretch}{1.1}
\caption{Effect of adding images (multimodal vs text-only) on QA performance, macro-averaged across models at the per-question level. For the p-value we used a paired sign-flip permutation test.}
\label{tab:multimodal_main_effect}

\resizebox{\columnwidth}{!}{
\begin{tabular}{|l|c|c|c|c|c|}
\hline
\textbf{Metric} & \textbf{Baseline} & \textbf{Mean $\Delta$} & \textbf{95\% CI} & \textbf{$p$-value} \\
\hline
F1        & 75.14\%  & +1.85 & [1.30, 2.42] & $< 10^{-4}$ \\
\hline
Precision & 77.58\% & +2.89 & [2.31, 3.51] & $< 10^{-4}$ \\
\hline
Recall    & 79.87\% & +0.63 & [0.06, 1.21] & 0.032 \\
\hline
\end{tabular}
}
\end{table}

We first evaluate whether incorporating visual context improves question answering performance when averaged across a set of large language models. For each question, we compute the mean performance across models under both text-only and multimodal conditions, and estimate paired differences using a sign-flip permutation test with bootstrap confidence intervals.
Table~\ref{tab:multimodal_main_effect} shows that multimodal inputs consistently improve performance across all evaluation metrics. We observe an average improvement of $+1.85$ F1 points (95\% CI: [1.30, 2.42], $p < 0.001$), indicating a statistically significant gain when visual context is available.

Improvements are particularly strong for precision ($+2.89$ points, $p < 0.001$), suggesting that visual information helps models reduce incorrect predictions. Recall shows a smaller but still statistically significant improvement ($+0.63$ points, $p = 0.032$), indicating a more modest effect on coverage.
These results demonstrate that adding images provides a consistent performance benefit across models and questions, with the strongest gains observed in precision.

\subsection{Effects of question complexity}

\begin{table}[t]
\centering
\small
\setlength{\tabcolsep}{4pt}
\renewcommand{\arraystretch}{1.1}
\caption{Baseline text-only performance and multimodal gain ($\Delta$ percentage points) across graph complexity tiers defined by triple count $n$.}
\label{tab:complexity_effect}

\resizebox{\columnwidth}{!}{
\begin{tabular}{l|cc|cc|cc}
\toprule
& \multicolumn{2}{c|}{\textbf{F1 (\%)}} & \multicolumn{2}{c|}{\textbf{Precision (\%)}} & \multicolumn{2}{c}{\textbf{Recall (\%)}} \\
\textbf{Complexity Tier} & \textbf{Base} & $\bm{\Delta}$ & \textbf{Base} & $\bm{\Delta}$ & \textbf{Base} & $\bm{\Delta}$ \\
\midrule
Low ($n \le 7$)         & 76.82 & \textbf{+4.65} & 79.10 & \textbf{+5.58} & 81.24 & \textbf{+2.86} \\
Medium ($7 < n \le 25$) & 75.11 & +0.47          & 77.62 & +0.96          & 79.80 & -0.22 \\
High ($25 < n$)         & 73.49 & +0.28          & 76.02 & +2.00          & 78.57 & -0.86 \\
\bottomrule
\end{tabular}
}
\end{table}

We analyze how the benefit of multimodal input varies with graph complexity.  Table~\ref{tab:complexity_effect} reports mean differences between multimodal  and text-only conditions, aggregated across models and questions within each  complexity tier. The results demonstrate that for low-complexity queries, adding  visual context yields a substantial improvement across all metrics, with a  +4.65 percentage-point gain in $F_1$ score ($p < 0.001$) and a +5.58 point gain  in precision. This indicates that multimodal input is particularly effective when  the reasoning structure is simple and direct. In contrast, the benefit decreases  sharply as graph complexity increases. For medium-complexity tasks, the net $F_1$  gain drops to a marginal +0.48 points ($\beta = -0.0417$ interaction term), while  recall slightly decreases (-0.22 points), suggesting that visual context provides  limited incremental value when reasoning chains lengthen. For high-complexity  questions, multimodal gains remain constrained: while precision shows a modest  improvement (+2.00 points), the overall net gain in $F_1$ is only +0.28 points  ($\beta = -0.0437$ interaction term), and recall declines further (-0.86 points),  pointing to potential visual noise or cognitive distraction effects in highly demanding  graph settings.

\begin{table}[t]
\centering
\small
\setlength{\tabcolsep}{6pt}
\renewcommand{\arraystretch}{1.2}
\caption{OLS regression results for $F_1$ score, modeling graph complexity tiers, prompt modality, and their interaction terms. Reference level is Low Complexity $\times$ Text-Only.}
\label{tab:ols_main}

\resizebox{\columnwidth}{!}{
\begin{tabular}{lccc}
\toprule
\textbf{Variable / Term} & \textbf{Coefficient ($\beta$)} & \textbf{95\% CI} & \textbf{$p$-value} \\
\midrule
Intercept (Low, Text-Only) & 0.8513 & [0.839, 0.863] & $< 0.001$ \\
\midrule
\multicolumn{4}{l}{\textit{Main Effects}} \\
Medium Complexity           & -0.0543 & [-0.072, -0.037] & $< 0.001$ \\
High Complexity             & -0.2504 & [-0.268, -0.233] & $< 0.001$ \\
Multimodal (Text + Image)   & +0.0465 & [0.029, 0.063]  & $< 0.001$ \\
\midrule
\multicolumn{4}{l}{\textit{Interaction Terms}} \\
Medium $\times$ Multimodal  & -0.0417 & [-0.066, -0.017] & 0.001 \\
High $\times$ Multimodal    & -0.0437 & [-0.068, -0.019] & $< 0.001$ \\
\bottomrule
\end{tabular}
}
\end{table}
\subsection{Regression Analysis}
\label{sec:regression_analysis}

To formally test how graph complexity, prompt modality, and their interaction jointly influence performance, we fit an Ordinary Least Squares (OLS) regression model with $F_1$ score as the dependent variable (Table~\ref{tab:ols_main}). 

The model confirms that structural complexity is a primary driver of performance degradation. Compared to the low-complexity baseline, medium-complexity subgraphs decrease $F_1$ by 0.0543 points ($p < 0.001$), while high-complexity subgraphs exhibit a severe drop of 0.2504 points ($p < 0.001$) regardless of prompt modality.

For low-complexity subgraphs, adding a graph visualization yields a significant positive main effect of $+0.0465$ $F_1$ points ($p < 0.001$). However, the negative interaction terms for medium complexity ($-0.0417$, $p = 0.001$) and high complexity ($-0.0437$, $p < 0.001$) almost entirely offset this advantage. The net estimated multimodal gain drops to $+0.0048$ $F_1$ points for medium-complexity and $+0.0028$ for high-complexity subgraphs. These empirical estimates closely match our non-parametric breakdown in Table~\ref{tab:complexity_effect}, confirming that visual prompting benefits are tightly bounded by subgraph scale and visual density.

\subsection{Model-wise effect of multimodal input}

\begin{figure}[t]
\centering
\includegraphics[width=\columnwidth]{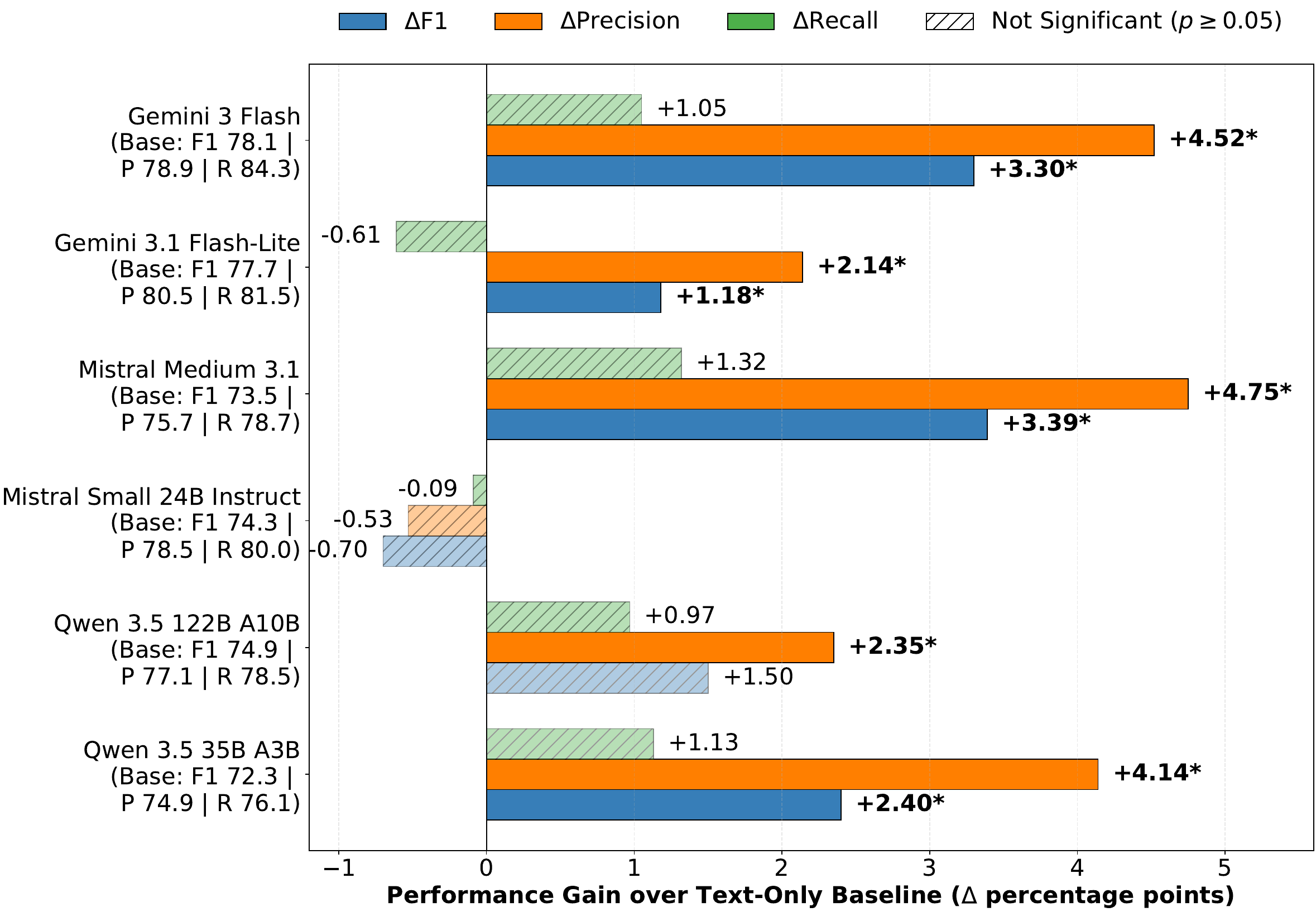}
\caption{Model-wise performance gains from multimodal input across all metrics. Significant at p < 0.05.}
\label{fig:modelwise}
\end{figure}

We examine how the benefit of multimodal input varies across different LLMs. Fig~\ref{fig:modelwise} reports the corresponding improvements of the multimodal setting over the text-only baseline.
Overall, multimodal input improves performance for most models, but the magnitude of the gain is highly model-dependent. The strongest and most consistent improvements are observed for \textbf{Gemini 3 Flash} and \textbf{Mistral Medium 3.1}, which achieve the largest gains in both F1 (+3.30 and +3.39, respectively) and precision (+4.52 and +4.75). These models also exhibit statistically significant improvements across all metrics, indicating a robust ability to leverage visual context.

In contrast, \textbf{Mistral Small 24B Instruct} shows no meaningful benefit from multimodal input, with small negative and non-significant changes across all metrics, suggesting limited capability to integrate visual information effectively at this scale.
Among the Qwen models, results are mixed. \textbf{Qwen 3.5 35B A3B} shows strong and statistically significant improvements in F1 (+2.40, $p < 0.001$) and precision (+4.14, $p < 10^{-7}$), while \textbf{Qwen 3.5 122B A10B} yields moderate gains that are primarily concentrated in precision rather than recall.

Across all models, the most consistent effect of multimodal input is an improvement in precision, whereas recall improvements are smaller and less stable, with occasional slight decreases.
Overall, \textbf{Gemini 3 Flash} and \textbf{Mistral Medium 3.1} emerge as the strongest beneficiaries of multimodal input in this setting in terms of absolute performance gain. In particular, \textbf{Gemini 3 Flash} with multimodal input achieves an F1 of 81.41 (+3.30 over text-only), precision of 83.39 (+4.52), and recall of 85.34 (+1.05), and across all scenarios appears to have the best performance in this task.

\subsection{Interaction between model size and multimodal input}
To examine whether the effect of multimodal input depends on model size, we estimated a linear regression model including an interaction between prompt type and model size. Model size was treated as a binary categorical variable, with models grouped into \textbf{small} and \textbf{medium} size regimes based on their relative scale within each model family, rather than exact parameter counts, which are not consistently disclosed across providers.

The interaction between prompt type and model size was positive but not statistically significant ($\beta = 0.0177$, $p = 0.10$). This indicates that there is no strong evidence that the effect of multimodal input differs between small and medium models.
Overall, the findings indicate that the benefits of multimodal prompting are broadly consistent across the two model sizes. While medium models appear to benefit slightly more from multimodal input than small models, the available evidence is insufficient to conclude that model size meaningfully moderates the effect of multimodality.

\subsection{Ablation Study of Visual Layout}

\begin{table}[t]
\centering
\small
\setlength{\tabcolsep}{8pt}
\renewcommand{\arraystretch}{1.15}
\caption{Ablation study of visual design choices and input modality conditions for Qwen 3.5 35B A3B. Values report absolute percentage point changes ($\Delta$) relative to its text-only baseline ($F_1 = 72.30\%$, $\text{Precision} = 74.90\%$, $\text{Recall} = 76.10\%$).}
\label{tab:ablation}

\resizebox{\columnwidth}{!}{
\begin{tabular}{lccc}
\toprule
\textbf{Condition} & $\bm{\Delta}$\textbf{F1} & $\bm{\Delta}$\textbf{Precision} & $\bm{\Delta}$\textbf{Recall} \\
\midrule
Image Only (Graphviz)           & -9.42 & -8.52 & -9.76 \\
Text + Image (Graphviz)         & +2.40 & +4.14 & +1.13 \\
Text + Image (Non-hierarchical) & +0.87 & +2.59 & -0.44 \\
Text + Image (NetworkX)         & -0.40 & +1.08 & -1.69 \\
Text + Reasoning-Path Images    & \textbf{+3.03} & \textbf{+3.90} & \textbf{+2.19} \\
\bottomrule
\end{tabular}
}
\end{table}

To better understand which aspects of the visual representation contribute to performance gains, we evaluated several alternative visualization strategies using Qwen 3.5 35B. Figures~\ref{fig:case1_graphs}, \ref{fig:case2_graphs}, and \ref{fig:case3_graphs} show examples of the different visualization modes, while Table~\ref{tab:ablation} reports their performance differences relative to the text-only baseline.

The strongest degradation is observed in the image-only condition, which reduces F1 by 9.42 percentage points. This result indicates that, for the evaluated model and setting, the graph image alone does not provide sufficient information for answering the questions and that the textual representation remains important.

Among multimodal configurations, the standard hierarchical visualization based on Graphviz improves F1 by 2.40 percentage points. When the same graph is rendered using a less structured layout, the gain drops to 0.87 points, suggesting that the organization of the visual representation affects how effectively the model can use the graph information.
Replacing the hierarchical Graphviz-based layout with a NetworkX-based layout eliminates the improvement entirely ($\Delta$F1=-0.40), further suggesting that the benefit does not arise merely from adding a graph image. Instead, the particular visual organization of the graph appears to matter.

The strongest result is obtained for the reasoning-path condition, which improves F1 by 3.03 percentage points over the text-only baseline. Because the underlying graph information is partitioned into separate path-specific images, this condition provides a more localized visual representation of the graph context. The result is consistent with our earlier complexity analysis, where the largest multimodal gains were observed for questions associated with smaller graph contexts. 

Overall, the ablation results suggest that multimodal gains depend not only on the presence of a graph image, but also on how the graph information is organized and presented. In particular, presenting graph information in structured, path-specific visual units appears promising, although the present ablation does not isolate the individual effects of image size, graph density, and reasoning-path structure.

\subsection{Qualitative Case Analysis}
\label{sec:case_analysis}

To complement the aggregate evaluation, we examine three examples using Qwen3.5-35B, illustrating multimodal improvement, multimodal degradation, and the effect of reasoning-path decomposition, respectively.

\paragraph{Multimodal Improvement via Visual Topological Filtering.}
To illustrate how multimodal input improves prediction precision, consider the question \emph{``What team that won the 2003 World Series championship did Pudge Rodriguez play for?''} (\textbf{WebQTest-1319}; Gold: \emph{Miami Marlins}). 
The retrieved evidence contains 12 reasoning paths due to Ivan Rodriguez's long career with the Texas Rangers, 6 paths terminate at \emph{Texas Rangers}, while only 2 terminate at \emph{Miami Marlins} alongside a single path connecting the \emph{2003 World Series} to the \emph{Miami Marlins}.
Under text-only prompting, the model suffers from frequency bias, predicting the statistically dominant entity \emph{Texas Rangers} 
In contrast, introducing graph visualizations (\emph{Text + Image}) corrects the prediction to \emph{Miami Marlins} across all evaluated layout renderings (e.g., Graphviz, NetworkX, and path decomposition). 
Apparently. visual presentation layout enforces topological path tracing: the VLM visually identifies \emph{Miami Marlins} as the sole converging sink for both query entities (\emph{Iv\'an Rodr\'iguez} and \emph{2003 World Series}), allowing the model to prune majority-class text distractors and satisfy the championship constraint. The analysis is detailed in Appendix~\ref{sec:case1_analysis}.

\paragraph{Multimodal Degradation via Visual Clutter.}
Visual context does not uniformly improve performance; dense or unorganized visual layouts can introduce perceptual errors even when the textual evidence remains identical. For the query \emph{``Whom was the person (that was born in Garfield) that Judy Garland was married to?''} (\textbf{WebQTrn-428}; Gold: \emph{Mickey Deans}), the text-only baseline correctly intersects the spouse and birthplace constraints. 
However, introducing full-graph visual prompting (\emph{Text + Image}) causes models to fail across multiple layout algorithms, predicting visually prominent distractors like \emph{James A. Garfield} or dumping all birthplace candidates (e.g., \emph{Stephen Fabian}, \emph{Tippy Larkin}) 
As detailed in Appendix~\ref{sec:case2_analysis}, we suppose this degradation stems from high-degree hub nodes (e.g., nationality nodes connecting dozens of irrelevant entities) that create dense radial clusters. The resulting visual clutter obscures localized target paths and induces lexical distraction around visually adjacent entity labels. This failure mode underscores the necessity of adaptive visual prompting policies that selectively suppress full-graph diagrams when subgraphs exhibit high visual density.

\paragraph{Reasoning-Path Decomposition and Visual Scalability.}
When retrieved subgraphs grow in scale and complexity, single-diagram visual representations can suffer from layout bottlenecks that degrade precision. For the query \emph{``What political district with governmental body European Parliament is where portuguese is spoken?''} (\textbf{WebQTrn-2496}; Gold: \emph{Portugal}), the retrieved evidence spans 51 triples and 28 reasoning paths. 
Both the text-only baseline and the hierarchical full-graph visualization (\emph{Graphviz}) suffer from distractor overload, returning 10--12 regional districts alongside \emph{Portugal} due to a failure to enforce joint constraint intersection. 
As detailed in Appendix~\ref{sec:case3_analysis}, while alternative spatial layouts (e.g., \emph{NetworkX}) mitigate this bottleneck by visually isolating clusters, decomposing the full graph into isolated reasoning-path images (\emph{Text + Path Images}) consistently achieves 100\% precision (\emph{['Portugal']}). 
By presenting distinct horizontal chains, path decomposition eliminates overlapping visual hairballs, allowing the VLM to sequentially trace intersecting constraints.

\vspace{0.3cm}

Together, these cases demonstrate that the effect of graph visualization is highly heterogeneous at the per-question level: visual prompting can prune text-frequency distractors (Case 1), introduce layout noise in dense radial clusters (Case 2), or require spatial layout restructuring and path decomposition to scale to complex subgraphs (Case 3). These qualitative insights directly complement our aggregate findings and suggest the need for adaptive visual prompting policies.

\section{Discussion}

Our results show that graph visualizations are beneficial for KGQA, but not unconditionally. Across models and instances, adding a graph image to the textual prompt improved F1 by 1.85 percentage points and precision by 2.89 points. The larger effect on precision suggests that the image helps models constrain their predictions and prune unsupported candidates -- a mechanism corroborated by our qualitative case analyzes (Appendix \ref{sec:case1_analysis}), rather than substantially increasing the amount of relevant information they recall. Thus, the visualization should be interpreted as complementary evidence that supports selection and verification during answer generation, rather than as a replacement for the textual representation ofa  KG.
The benefit of a visualization appears to depend on the complexity of the retrieved graph context. For low-complexity instances, adding an image improved F1 by 4.65 points, whereas the improvement was close to zero for medium- and high-complexity instances. The regression analysis is consistent with this pattern, showing that the positive effect of multimodal prompting decreases as complexity increases. One possible explanation is that visualizations are more useful when the relevant subgraph is sufficiently small for its topology, local neighborhoods, and relation paths to remain visually accessible. For larger graphs, a single image may be more difficult to process and may provide less useful structural information.

This result can be interpreted through Moody's \textit{Physics of Notations} \citep{moody2009_notation}, which identifies principles for cognitively effective visual notations, including perceptual discriminability, complexity management, and cognitive integration. Although developed for human users, this framework provides a vocabulary for comparing how visual design affects VLM performance. We do not assume that VLMs process diagrams like humans; rather, our findings offer preliminary evidence toward a VLM-oriented account of which visual properties support multimodal prompting.

In our setting, the hierarchical Graphviz diagrams appear to improve the perceptual accessibility of the relational structure: nodes, directed edges, and their local arrangement provide cues that are not directly available in a linear list of triples. This can be interpreted using Paivio's principle of \textit{dual coding} \citep{paivio1990_dual_coding} mentioned by Moody, according to which text and graphics are most effective when they convey complementary and mutually reinforcing information. In this case, this principle appears to transfer to VLMs in the KGQA task, where the multimodal condition outperforms both the text-only and image-only conditions.
At the same time, the ablation results demonstrate that the presence of an image alone is insufficient. Removing textual triples and using only a Graphviz image reduces F1 by 9.42 points relative to the text-only baseline. This indicates that the visual representation does not preserve all task-relevant information in a form that current VLMs can reliably decode. While Moody argues that for human end users visual notations tend to convey information more effectively than text, we can see that for VLMs in the KGQA task, text remains necessary for explicit relation semantics, entity names, and precise answer grounding. In this case, rather than treating the graph image as a substitute for the KG, it is more appropriate to treat it as a secondary representation that exposes structural regularities while the text maintains semantic precision.

Following \cite{cheng2004_sometimes} and \cite{moody2009_notation}, our comparison between rendering strategies highlights the critical role of visual design in multimodal VLM prompting. Across the aggregate benchmark, a hierarchical Graphviz layout yields a 2.40-point F1 improvement, whereas an unstructured layout reduces this gain to 0.87 points, and the NetworkX layout shows no overall improvement over text alone. However, case-level analyses reveal nuance behind these averages. These findings demonstrate that visual layout is not merely cosmetic—it determines whether a rendering exposes or obscures essential relational topology. 
These differences support, in the case of VLMs, Moody's claim that visual syntax affects cognitive effectiveness independently of semantic content. All variants encode the same underlying graph but differ in their ability to manage visual complexity and make the relevant structure discriminable. In particular, we gather evidence that for VLMs the layout is not merely a cosmetic property: it determines whether the visual representation exposes or obscures the graph structure that the model must use.

 Our findings fall within a broader literature on visual understanding in VLMs. Recent benchmarks on charts, documents, and infographics show that VLMs can extract high-level patterns and produce fluent descriptions, but still struggle with fine-grained quantitative reasoning, spatial relations, and factual faithfulness, especially as visual complexity increases \citep[e.g.,][]{islam2024_chart_vlms,huang2024_chocolate}. In many of these settings, models rely heavily on textual elements (titles, legends, annotations) and their performance is highly sensitive to layout quality and visual clutter. Our results mirror this general pattern: pretrained VLMs derive measurable benefit from an additional visual channel when the diagram is relatively simple and well-structured, but the gains diminish or disappear for complex graphs and visually dense renderings. This observation is consistent with graph-specific studies such as GraphVis \citep{deng_2025graphvis} and GITA \citep{wei2024_GITA}, which report significant improvements only after graph-tailored visual training and careful layout control, and with surveys of VLMs that emphasize the need for better compositional visual reasoning and robust handling of complex diagrams \citep{ke2025_compositional_visual_reasoning,li2025_lvml_survey}.

\section{Conclusion}

Overall, the practical implication of our results is that KG visualizations should be selected adaptively rather than appended to every prompt to improve performance in the KGQA task. A suitable policy would add a diagram for small or moderately sized, visually sparse subgraphs; prefer hierarchical layouts with clear directionality and readable labels; retain textual triples in all cases; and decompose dense contexts into path-level or other focused visualizations. For large or highly complex queries, the expected value of a single full-graph image is limited and may even be negative. We hypothesize that adaptive visual prompting strategies can improve scores on KGQA and, more broadly, on visual reasoning benchmarks.

Another important research direction is to gather more evidence connecting these empirical findings to principled theories of visual notation design, such as Moody’s Physics of Notations. Future work could systematically evaluate different structured visualization styles against these principles, treating VLM prompts as visual notations, and asking which combinations of graphical and textual encoding maximize effectiveness for multimodal models. This would move beyond ad hoc rendering choices towards a principled integration of visually structured data into prompts.

More broadly, we see our work as a first step toward a design discipline for multimodal prompting with structured data. Extending the analysis to other diagram types (tables, flowcharts, multi-chart dashboards, multi-graph layouts) would allow direct comparison of VLM performance across notations that differ in how they instantiate visualization principles. Additional research should also consider how easily diagrams can be generated automatically from structured data, and how visual dialects might be tuned to different tasks or model families. Ultimately, a theory-driven account of when and how to visualize structured information could inform both the design of future VLM architectures, including effective multimodal KGQA systems.

\section*{Limitations}

This study focuses on KGQA on a specific set of datasets, one specific KGQA pipeline (ROG) and 6 VLM models, so the observed multimodal gains may not transfer directly to other tasks, domains, or architectures. Our notion of complexity is also approximate and tied to one type of graph context, and we explore only a small slice of the visual design space (a few layouts and decomposition strategies).
Moreover, we analyze performance using standard automatic metrics and a fixed prompting setup, without human evaluations of interpretability or trust, and without considering how training-time exposure to diagrams might interact with inference-time prompting. These constraints mean that our recommendations on when and how to add KG visualizations should be viewed as preliminary and task-specific, motivating future work that systematically varies notation design, model training, and evaluation regimes.

Furthermore, we have not explicitly addressed the additional computational and API costs associated with multimodal prompting. Compared with text-only prompts, graph images may introduce additional input processing, image-token or vision-processing costs, increased latency, and infrastructure requirements, depending on the model and serving platform. Given the relatively modest average performance improvement observed in our experiments, the practical benefit of adding a visualization may, therefore, depend on the specific application and its cost constraints. 

\section*{Acknowledgments}
This publication was funded by a flagship project ``CHExRISH: Cultural Heritage Exploration and Retrieval with Intelligent Systems at Jagiellonian University'' under the Strategic Programme Excellence Initiative at Jagiellonian University.

The research for this publication has been supported by a grant from the Priority Research Area DigiWorld under the Strategic Programme Excellence Initiative at Jagiellonian University.

This research is supported by SFI MediaFutures partners and the Research Council of Norway (grant number 309339).

During the preparation of this work, the authors used GPT-5.1 and GPT-4o in order to: grammar and spelling check, paraphrase and reword. After using these services, the authors reviewed and edited the content as needed and take full responsibility for the publication’s content.

\bibliography{custom}

\newpage
\appendix
\section{Prompt Template}
\label{app:prompts}

We used a shared prompt template across experimental conditions. The
\texttt{\{type\}} placeholder was instantiated as either \emph{reasoning
paths}, \emph{reasoning paths and the image}, or \emph{image}, depending on
the input condition. The placeholders \texttt{\{reasoning\_paths\}} and
\texttt{\{question\}} were replaced with the retrieved reasoning paths and
the input question, respectively. In the image-only condition, the
\textit{Reasoning Paths} field was omitted.

\begin{lstlisting}[basicstyle=\ttfamily\footnotesize, breaklines=true, frame=single]
Use the {type} to answer the question.

Output format:
["answer1", "answer2", ...]

Rules:
- Return ONLY a Python list of answers.
- Do not provide explanations, reasoning, or additional text.
- Do not restate the question.
- Include all possible correct answers.

Reasoning Paths:
{reasoning_paths}

Question:
{question}

Answer:
\end{lstlisting}

\section{Methodology: Evidence Construction \& Graph Layouts}
\label{sec:methodology_construction}

To isolate the contribution of visual graph representations during answer generation, our evaluation fixes the retrieved Knowledge Graph (KG) context between textual and multimodal conditions.

\subsection{From RoG Paths to Prompts and Visual Renderings}
Following the Reasoning on Graphs (RoG) pipeline \citep{luo_reasoning_2024}, evidence retrieval and prompt generation proceed through three main steps:

\paragraph{1. Reasoning Path Extraction.} 
Given a question $q$, relation paths are grounded against Freebase to retrieve sequence chains $\mathcal{W}_z = \{e_0 \xrightarrow{r_1} e_1 \xrightarrow{r_2} \dots \xrightarrow{r_k} e_k\}$, where $e_0 \in \mathcal{E}_q$ represents a question entity, $r_i \in \mathcal{R}$ denotes a relation, and $e_k$ is a candidate answer entity.

\paragraph{2. Conversion to Textual Context.}
The retrieved paths $\mathcal{W}_z$ are merged and deduplicated into a set of distinct directed triples $\mathcal{T} = \{(e_s, r, e_o)\}$. For text-only prompts, these triples are linearized sequentially into structured natural language strings using the format:
\begin{center}
\texttt{[Entity 1]} $\rightarrow$ \texttt{[Relation]} $\rightarrow$ \texttt{[Entity 2]}
\end{center}

\paragraph{3. Visual Graph Rendering.}
The triple set $\mathcal{T}$ is compiled into a Graphviz specification $\mathcal{G} = (\mathcal{V}, \mathcal{E})$ and rendered into visual prompt images under different structural policies:
\begin{itemize}
    \item \textbf{Hierarchical Layout (\texttt{graphviz}):} Rendered using the directed acyclic graph (\texttt{dot}) engine with left-to-right orientation to emphasize directional flow.
    \item \textbf{Unstructured Layout (\texttt{graphviz\_chaos}):} Disables hierarchical constraints, positioning nodes without spatial directional layout.
    \item \textbf{Force-Directed Layout (\texttt{networkx}):} Rendered via NetworkX's \texttt{spring\_layout}, causing high-degree entities to cluster radially.
    \item \textbf{Path Decomposition (\texttt{reasoning\_paths}):} Each path $w_z \in \mathcal{W}_z$ is rendered as an independent, single-chain diagram. The VLM receives the textual context alongside an array of these separated path images.
\end{itemize}

\section{Case Study 1: Visual Topological Filtering (WebQTest-1319)}
\label{sec:case1_analysis}

\begin{table*}[h!]
\small
\centering
\begin{tabular}{lp{10.2cm}}
\toprule
\textbf{Dataset ID} & \texttt{WebQTest-1319\_448df18f183f1782229b1d1945aad5ac} (CWQ) \\
\textbf{Question} & What team that won the 2003 World Series championship did Pudge Rodriguez play for? \\
\textbf{Gold Answer} & \texttt{['Miami Marlins']} \\
\midrule
\textbf{Reasoning paths} & 
1. \texttt{Iván Rodríguez} $\rightarrow$ \texttt{sports.sports\_team\_roster.player} $\rightarrow$ \texttt{m.0z9w064} $\rightarrow$ \texttt{Miami Marlins} \\
& 2. \texttt{Iván Rodríguez} $\rightarrow$ \texttt{sports.sports\_team\_roster.player} $\rightarrow$ \texttt{m.0ywy2gy} $\rightarrow$ \texttt{Texas Rangers} \\
& 3--6, 11. \texttt{Iván Rodríguez} $\rightarrow$ \texttt{baseball.batting\_statistics} (5 paths) $\rightarrow$ \texttt{Texas Rangers} \\
& 7. \texttt{Iván Rodríguez} $\rightarrow$ \texttt{baseball.batting\_statistics.player} $\rightarrow$ \texttt{m.06sbxjs} $\rightarrow$ \texttt{Miami Marlins} \\
& 8--9. \texttt{Iván Rodríguez} $\rightarrow$ \texttt{baseball.batting\_statistics} (2 paths) $\rightarrow$ \texttt{Detroit Tigers} \\
& 10. \texttt{Iván Rodríguez} $\rightarrow$ \texttt{baseball.batting\_statistics} (1 path) $\rightarrow$ \texttt{New York Yankees} \\
& 12. \texttt{2003 World Series} $\rightarrow$ \texttt{sports.sports\_team.championships} $\rightarrow$ \texttt{Miami Marlins} \\
\midrule
\textbf{Retrieved Triples} & 
1. \texttt{(m.0z9w064, sports.sports\_team\_roster.team, Miami Marlins)} \\
& 2. \texttt{(m.06sbxd2, baseball.batting\_statistics.team, Texas Rangers)} \\
& 3. \texttt{(Iván Rodríguez, sports.sports\_team\_roster.player, m.0z9w064)} \\
& 4. \texttt{(2003 World Series, sports.sports\_team.championships, Miami Marlins)} \\
& 5. \texttt{(m.06sbxjs, baseball.batting\_statistics.team, Miami Marlins)} \\
& 6. \texttt{(Iván Rodríguez, baseball.batting\_statistics.player, m.06sbxh6)} \\
& 7. \texttt{(Iván Rodríguez, baseball.batting\_statistics.player, m.06sbxkk)} \\
& 8. \texttt{(Iván Rodríguez, baseball.batting\_statistics.player, m.06sbxpg)} \\
& 9. \texttt{(Iván Rodríguez, baseball.batting\_statistics.player, m.06sbxh\_)} \\
& 10. \texttt{(m.0ywy2gy, sports.sports\_team\_roster.team, Texas Rangers)} \\
& 11. \texttt{(m.06sbxh\_, baseball.batting\_statistics.team, Texas Rangers)} \\
& 12. \texttt{(Iván Rodríguez, baseball.batting\_statistics.player, m.06sbxlb)} \\
& 13. \texttt{(m.06sbxkk, baseball.batting\_statistics.team, Detroit Tigers)} \\
& 14. \texttt{(m.06sbxlb, baseball.batting\_statistics.team, Detroit Tigers)} \\
& 15. \texttt{(Iván Rodríguez, baseball.batting\_statistics.player, m.06sbxbj)} \\
& 16. \texttt{(Iván Rodríguez, sports.sports\_team\_roster.player, m.0ywy2gy)} \\
& 17. \texttt{(m.06sbxfn, baseball.batting\_statistics.team, Texas Rangers)} \\
& 18. \texttt{(m.06sbxpg, baseball.batting\_statistics.team, New York Yankees)} \\
& 19. \texttt{(Iván Rodríguez, baseball.batting\_statistics.player, m.06sbxjs)} \\
& 20. \texttt{(Iván Rodríguez, baseball.batting\_statistics.player, m.06sbxd2)} \\
& 21. \texttt{(Iván Rodríguez, baseball.batting\_statistics.player, m.06sbxfn)} \\
& 22. \texttt{(m.06sbxh6, baseball.batting\_statistics.team, Texas Rangers)} \\
& 23. \texttt{(m.06sbxbj, baseball.batting\_statistics.team, Texas Rangers)} \\
\midrule
\textbf{Model outputs for Qwen3.5-35b-a3b} & \\
-- \emph{Text-Only} & \texttt{['Texas Rangers']} \quad \textcolor{red}{$\bm{\times}$ (Majority-Class Text Frequency Bias)} \\
-- \emph{Image-Only} & \texttt{['Florida Marlins']} \quad \textcolor{orange}{$\bm{\sim}$ (Name Variance without Text Grounding)} \\
-- \emph{Text + Image (non-hierarchical Graphviz)} & \texttt{['Miami Marlins']} \quad \textcolor{teal}{$\bm{\checkmark}$ (Correct)} \\
-- \emph{Text + Image (NetworkX)} & \texttt{['Miami Marlins']} \quad \textcolor{teal}{$\bm{\checkmark}$ (Correct)} \\
-- \emph{Text + Image (Graphviz)} & \texttt{['Miami Marlins']} \quad \textcolor{teal}{$\bm{\checkmark}$ (Correct)} \\
-- \emph{Text + Path Images} & \texttt{['Miami Marlins']} \quad \textcolor{teal}{$\bm{\checkmark}$ (Correct)} \\
\bottomrule
\end{tabular}
\caption{\textbf{Matched Analysis for Case 1 (WebQTest-1319).} Comparison showing how visual topology allows the model to prune majority-class textual frequency distractors.}
\label{tab:case1_details}
\end{table*}

Case 1 (\autoref{tab:case1_details}) demonstrates how visual graph prompting overcomes token frequency distraction in textual contexts. Figure~\ref{fig:case1_graphs} presents the different visualization modes for this graph context. From this case, we can point to the following mechanisms:

\begin{itemize}
    \item \textbf{Text-Only Frequency Bias:} Because Iván Rodríguez spent a major portion of his career with the Texas Rangers, 6 of the 12 retrieved paths terminate at \texttt{Texas Rangers}. In contrast, only 2 paths lead to \texttt{Miami Marlins}, alongside a single path connecting \texttt{2003 World Series} to \texttt{Miami Marlins}. Under text-only prompting, the VLM over-indexes on the high string frequency of \texttt{Texas Rangers}, outputting the wrong team.
    \item \textbf{Topological Pruning via Visual Prompting:} When visual context is added (\emph{Text + Image}), the model correctly predicts \texttt{Miami Marlins} across all visual layout variants (\texttt{graphviz}, \texttt{graphviz\_chaos}, \texttt{networkx}, and \texttt{reasoning\_paths}). Spatially laying out nodes forces path tracing over token frequency. The VLM visually tracks \texttt{Miami Marlins} as the single intersecting sink node connected to both query entities (\texttt{Iván Rodríguez} and \texttt{2003 World Series}), pruning out the non-intersecting \texttt{Texas Rangers} majority cluster.
    \item \textbf{Text-Visual Synergy (Image-Only Behavior):} Under \emph{Image-Only} prompting, the model outputs \texttt{Florida Marlins}. While historically accurate for the 2003 World Series, it fails to match the exact KB entity string (\texttt{Miami Marlins}) present in the textual evidence. This confirms that visual context acts as a spatial constraint filter over textual facts rather than replacing textual grounding altogether.
\end{itemize}

\section{Case Study 2: Visual Distraction in High-Fanout Subgraphs (WebQTrn-428)}
\label{sec:case2_analysis}

\begin{table*}[h!]
\small
\centering
\begin{tabular}{lp{10.2cm}}
\toprule
\textbf{Dataset ID} & \texttt{WebQTrn-428\_014f6ffa694535e10058379d5281b3ee} (CWQ) \\
\textbf{Question} & Whom was the person (that was born in Garfield) that Judy Garland was married to? \\
\textbf{Gold Answer} & \texttt{['Mickey Deans']} \\
\midrule
\textbf{Reasoning Paths} & 
1. \texttt{Garfield} $\rightarrow$ \texttt{people\_born\_here} $\rightarrow$ \texttt{Mickey Deans} \\
& 2. \texttt{Garfield} $\rightarrow$ \texttt{people\_born\_here} $\rightarrow$ \texttt{Tippy Larkin} \\
& 3. \texttt{Garfield} $\rightarrow$ \texttt{people\_born\_here} $\rightarrow$ \texttt{Stephen Fabian} \\
& 4. \texttt{Marriage} $\rightarrow$ \texttt{type\_of\_union} $\rightarrow$ \texttt{m.0j4k8cq} $\rightarrow$ \texttt{spouse\_s} $\rightarrow$ \texttt{James A. Garfield} \\
& 5. \texttt{Marriage} $\rightarrow$ \texttt{type\_of\_union} $\rightarrow$ \texttt{m.0wqk20x} $\rightarrow$ \texttt{spouse\_s} $\rightarrow$ \texttt{Judy Garland} \\
& 6. \texttt{Marriage} $\rightarrow$ \texttt{type\_of\_union} $\rightarrow$ \texttt{m.0hj5pf2} $\rightarrow$ \texttt{spouse\_s} $\rightarrow$ \texttt{Linda Bove} \\
& 7. \texttt{Marriage} $\rightarrow$ \texttt{type\_of\_union} $\rightarrow$ \texttt{m.02kkl\_c} $\rightarrow$ \texttt{spouse\_s} $\rightarrow$ \texttt{Mark Herron} \\
& 8. \texttt{Marriage} $\rightarrow$ \texttt{type\_of\_union} $\rightarrow$ \texttt{m.02kkltn} $\rightarrow$ \texttt{spouse\_s} $\rightarrow$ \texttt{Sidney Luft} \\
& 9. \texttt{Judy Garland} $\rightarrow$ \texttt{nationality} $\rightarrow$ \texttt{USA} $\rightarrow$ \texttt{nationality} $\rightarrow$ \texttt{Mickey Deans} \\
& 10--34. \texttt{Judy Garland} $\rightarrow$ \texttt{nationality} $\rightarrow$ \texttt{USA} $\rightarrow$ \texttt{nationality} $\rightarrow$ \texttt{[20+ Distractor Entities]} \\
\midrule
\textbf{Retrieved Triples} & 
1. \texttt{(Garfield, location.location.people\_born\_here, Mickey Deans)} \\
& 2. \texttt{(Garfield, location.location.people\_born\_here, Tippy Larkin)} \\
& 3. \texttt{(Garfield, location.location.people\_born\_here, Stephen Fabian)} \\
& 4. \texttt{(Judy Garland, location.location.people\_born\_here, Grand Rapids)} \\
& 5. \texttt{(Marriage, type\_of\_union, m.02kkltn)} $\rightarrow$ \texttt{(m.02kkltn, spouse\_s, Sidney Luft)} \\
& 6. \texttt{(Marriage, type\_of\_union, m.02kkl\_c)} $\rightarrow$ \texttt{(m.02kkl\_c, spouse\_s, Mark Herron)} \\
& 7. \texttt{(Marriage, type\_of\_union, m.0j4k8cq)} $\rightarrow$ \texttt{(m.0j4k8cq, spouse\_s, James A. Garfield)} \\
& 8. \texttt{(Marriage, type\_of\_union, m.0wqk20x)} $\rightarrow$ \texttt{(m.0wqk20x, spouse\_s, Judy Garland)} \\
& 9. \texttt{(Marriage, type\_of\_union, m.0hj5pf2)} $\rightarrow$ \texttt{(m.0hj5pf2, spouse\_s, Linda Bove)} \\
& 10--28. \texttt{(Judy Garland, nationality, USA)} $\rightarrow$ \texttt{(USA, nationality, [20+ US Entities])} \\
\midrule
\multicolumn{2}{l}{\textbf{Model Outputs for Qwen3.5-35B-A3B}} \\
-- \emph{Text-Only} & \texttt{['Mickey Deans']} \quad \textcolor{teal}{$\bm{\checkmark}$ (Correct Constraint Intersection)} \\
-- \emph{Image-Only} & \texttt{['Harold James Green', 'Mickey Rooney', 'Vincente Minnelli', 'Sidney Luft', 'Mark Hellerman']} \quad \textcolor{red}{$\bm{\times}$ (Ungrounded Hallucinations)} \\
-- \emph{Text + Image (Graphviz)} & \texttt{['James A. Garfield']} \quad \textcolor{red}{$\bm{\times}$ (Entity Label Distraction)} \\
-- \emph{Text + Image (non-hierarchical Graphviz)} & \texttt{['Sidney Luft']} \quad \textcolor{red}{$\bm{\times}$ (Omitted Birthplace Constraint)} \\
-- \emph{Text + Image (NetworkX)} & \texttt{['Linda Bove']} \quad \textcolor{red}{$\bm{\times}$ (Layout Distortion Error)} \\
-- \emph{Text + Path Images} & \texttt{['Sidney Luft', 'Mickey Deans']} \quad \textcolor{orange}{$\bm{\sim}$ (Partial Noise Reduction)} \\
\bottomrule
\end{tabular}
\caption{\textbf{Matched Analysis for Case 2 (WebQTrn-428).} Example demonstrating visual prompting degradation due to hub-entity radial clutter, reasoning path noise, and visual entity-label distraction.}
\label{tab:case2_details}
\end{table*}

\begin{figure*}[t]
    \centering

    \begin{subfigure}[t]{0.43\textwidth}
        \centering
        \includegraphics[width=\linewidth,height=0.27\textheight,keepaspectratio]
        {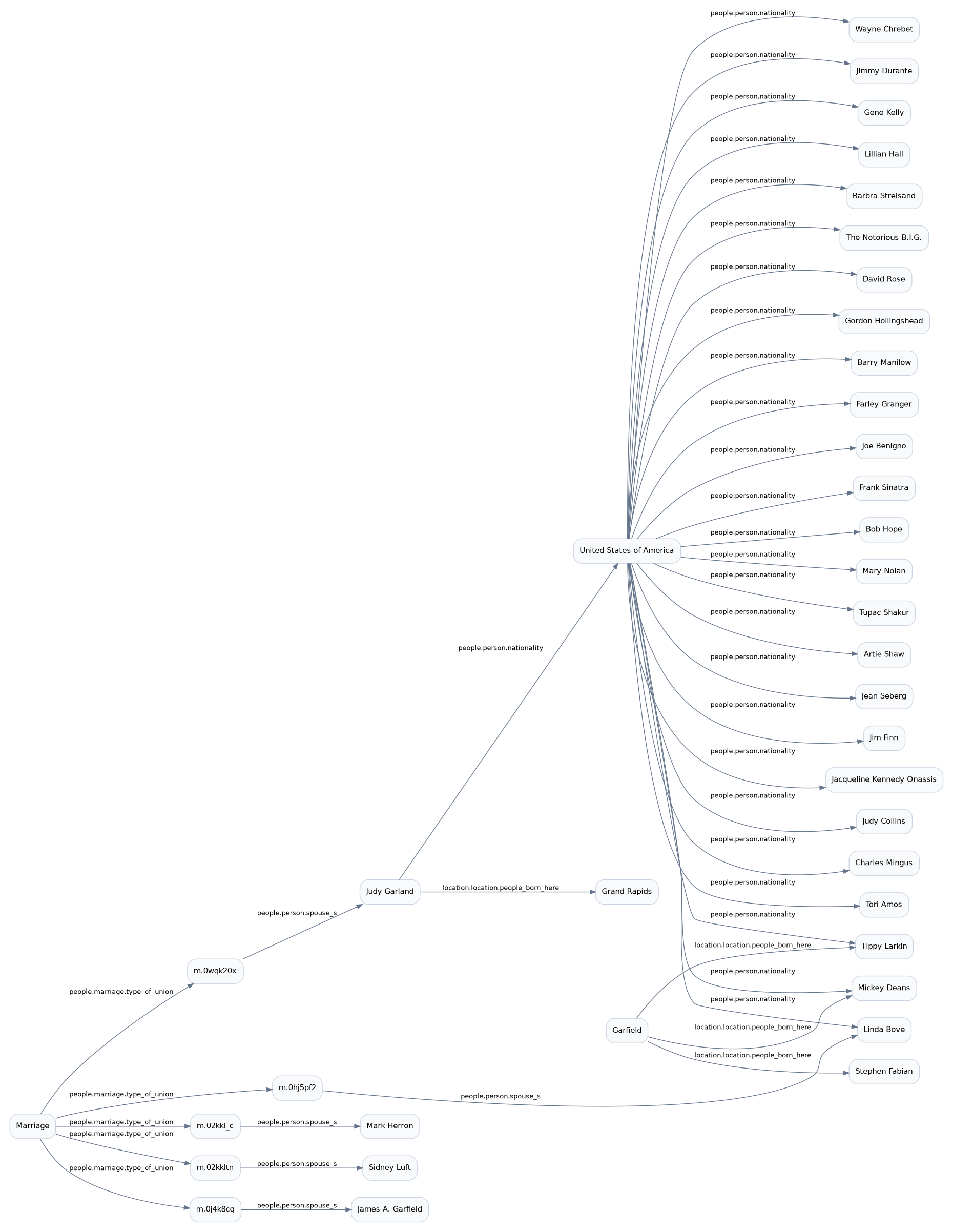}
        \caption{Graphviz}
        \label{fig:graphviz}
    \end{subfigure}
    \hfill
    \begin{subfigure}[t]{0.43\textwidth}
        \centering
        \includegraphics[width=\linewidth,height=0.27\textheight,keepaspectratio]
        {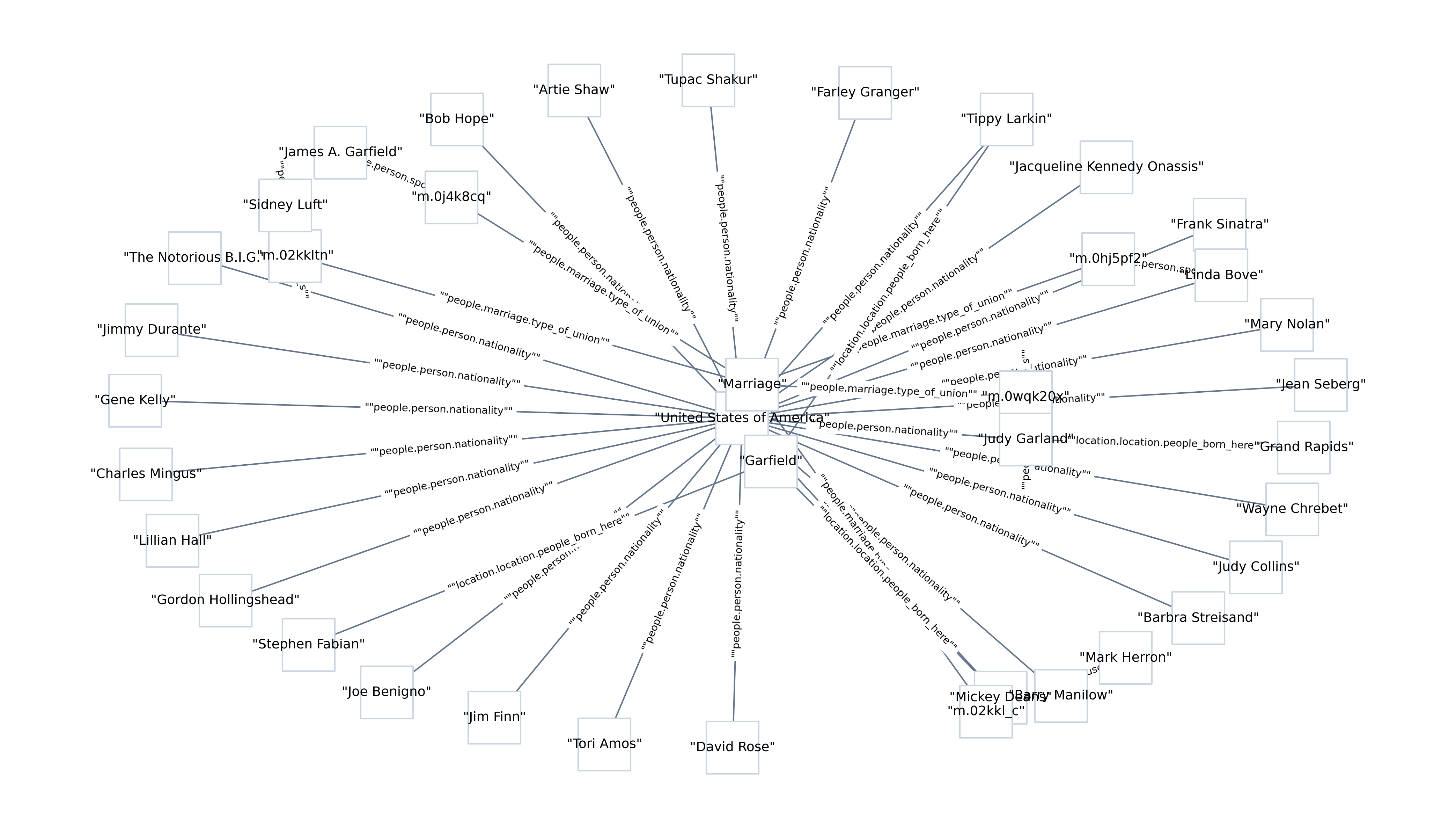}
        \caption{NetworkX}
        \label{fig:networkx}
    \end{subfigure}

    \vspace{0.5em}

    \begin{subfigure}[t]{0.43\textwidth}
        \centering
        \includegraphics[width=\linewidth,height=0.27\textheight,keepaspectratio]
        {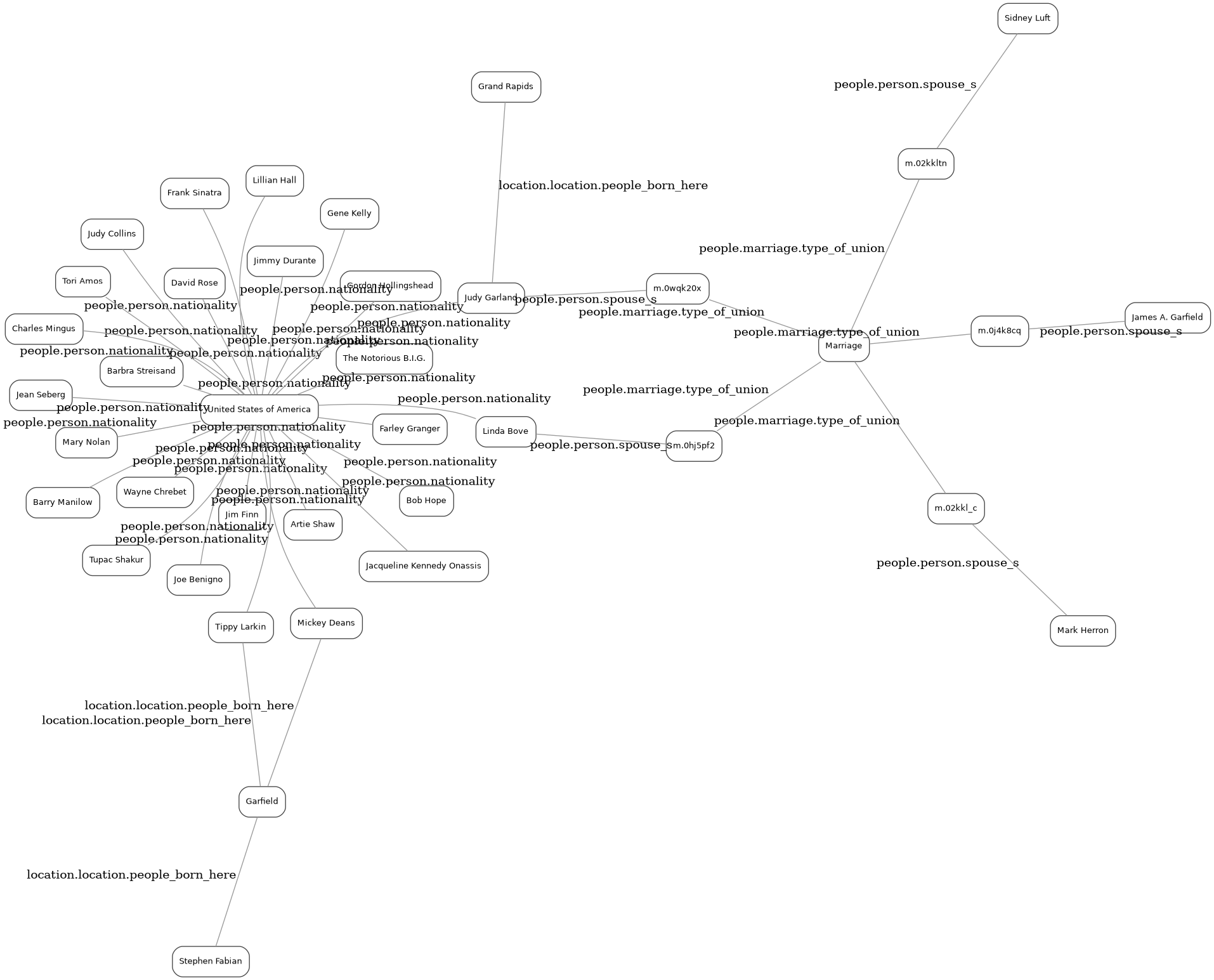}
        \caption{Graphviz with a non-hierarchical layout}
        \label{fig:graphviz-chaos}
    \end{subfigure}
    \hfill
    \begin{subfigure}[t]{0.43\textwidth}
        \centering
        \includegraphics[width=\linewidth,height=0.27\textheight,keepaspectratio]
        {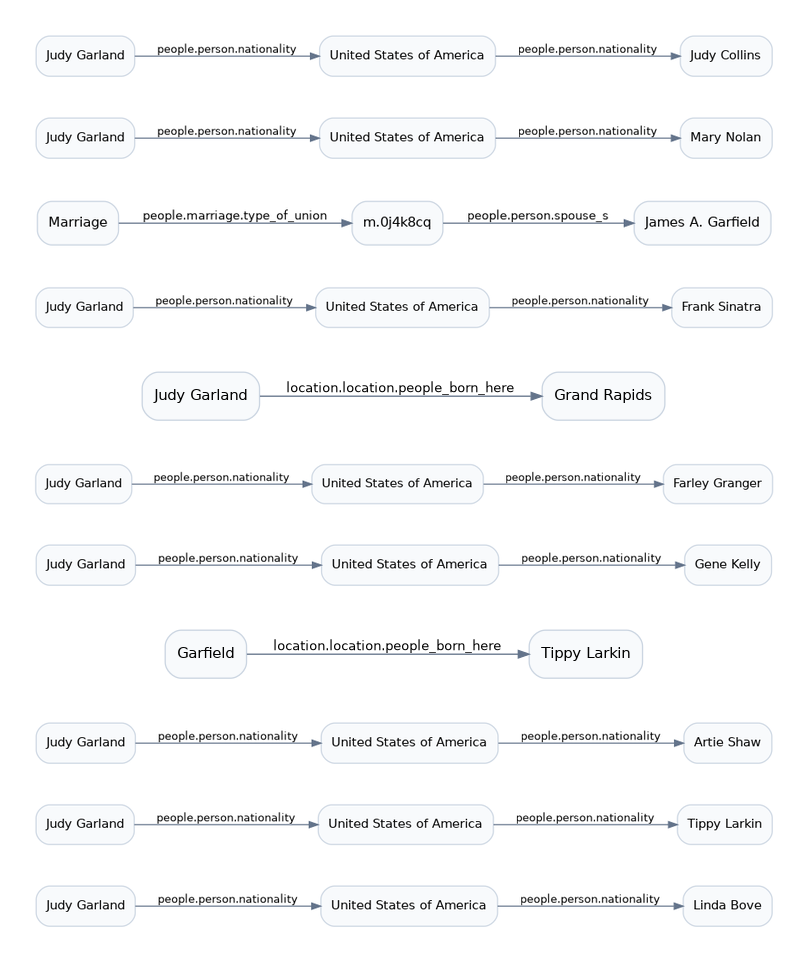}
        \caption{Excerpt of reasoning paths}
        \label{fig:reasoning-paths}
    \end{subfigure}

    \caption{The four visualizations for case 2 (\textbf{WebQTest-428}):
    (a) Graphviz, (b) NetworkX, (c) Graphviz with a non-hierarchical layout,
    and (d) excerpt of reasoning paths. }
    \label{fig:case2_graphs}
\end{figure*}

Case 2 (\autoref{tab:case2_details}) provides a critical failure analysis showing where multimodal prompting degrades performance compared to text-only baselines. Figure~\ref{fig:case2_graphs} presents the different visualization modes for this graph context. From this case, we can point to the following mechanisms:
\begin{itemize}
    \item \textbf{Text-Only Success:} Textual LLMs easily parse the linearized chains, intersecting \texttt{Judy Garland} $\rightarrow$ \texttt{spouse} $\rightarrow$ \texttt{Mickey Deans} with \texttt{Garfield} $\rightarrow$ \texttt{people\_born\_here} $\rightarrow$ \texttt{Mickey Deans}.
    \item \textbf{Visual Radial Clutter:} The node \texttt{United States of America} introduces 20+ irrelevantly connected entities (e.g., \emph{Tupac Shakur}, \emph{Bob Hope}), forming a massive visual star cluster that visually obscures the small 2-edge paths connecting \texttt{Garfield} and \texttt{Mickey Deans}.
    \item \textbf{Visual Entity Label Distraction:} In single-image visual prompts (\texttt{Text + Image}), models frequently over-index on the text label \texttt{James A. Garfield} due to lexical similarity with the question entity \texttt{Garfield}, erroneously selecting him as the spouse.
    \item \textbf{Adaptive Prompting Justification:} This case validates our adaptive policy: subgraphs containing high-degree hub entities or dense star topologies should suppress visual diagram generation to avoid degrading LLM reasoning.
\end{itemize}

\section{Case Study 3: High-Complexity Graph Scalability \& Layout Sensitivity (WebQTrn-2496)}
\label{sec:case3_analysis}

\begin{table*}[h!]
\small
\centering
\begin{tabular}{lp{10.2cm}}
\toprule
\textbf{Dataset ID} & \texttt{WebQTrn-2496\_d883a8a48671befac33bf89e3d801830} (CWQ) \\
\textbf{Question} & What political district with governmental body European Parliament is where portuguese is spoken? \\
\textbf{Gold Answer} & \texttt{['Portugal']} \\

\midrule
\textbf{Reasoning Paths} & 
1. \texttt{Portuguese Language} $\rightarrow$ \texttt{languages\_spoken} $\rightarrow$ \texttt{Portugal} \\
& 2. \texttt{European Parliament} $\rightarrow$ \texttt{governmental\_body.members} $\rightarrow$ \texttt{m.011ccr9x} $\rightarrow$ \texttt{representatives} $\rightarrow$ \texttt{Portugal} \\
& 3. \texttt{European Parliament} $\rightarrow$ \texttt{position\_held} $\rightarrow$ \texttt{m.063x9d1} $\rightarrow$ \texttt{district\_represented} $\rightarrow$ \texttt{Portugal} \\
& 4. \texttt{Portuguese Language} $\rightarrow$ \texttt{languages\_spoken} $\rightarrow$ \texttt{Cape Verde} / \texttt{Angola} / \texttt{Guinea-Bissau} / \texttt{Macau} (4 paths) \\
& 5--28. \texttt{European Parliament} $\rightarrow$ \texttt{[member/position]} $\rightarrow$ \texttt{[Estonia, Denmark, Latvia, UK Districts, etc.]} (22 paths) \\
\midrule
\textbf{Retrieved Triples} & 
1. \texttt{(Portuguese Language, location.country.languages\_spoken, Portugal)} \\
& 2. \texttt{(m.011ccr9x, government.political\_district.representatives, Portugal)} \\
& 3. \texttt{(European Parliament, government.governmental\_body.members, m.011ccr9x)} \\
& 4. \texttt{(m.09tphf2, government.political\_district.representatives, Estonia)} \\
& 5. \texttt{(Portuguese Language, location.country.languages\_spoken, Cape Verde)} \\
& 6--51. \texttt{(European Parliament, government.governmental\_body.members, [20+ regional member MIDs])} \\
\midrule
\midrule

\multicolumn{2}{l}{\textbf{Model Outputs for Qwen3.5-35B-A3B}} \\
-- \emph{Text-Only} & \texttt{['Portugal', 'Estonia', 'Latvia', 'Denmark', 'North West England', 'Yorkshire and the Humber', 'North-West', 'North-West France', 'West Midlands', 'South East England', 'South', 'South-East England']} \quad \textcolor{red}{$\bm{\times}$ (Low Precision / Distractor Overload)} \\
-- \emph{Image-Only} & \texttt{['Azores', 'Madeira']} \quad \textcolor{red}{$\bm{\times}$ (Ungrounded Regional Hallucination)} \\
-- \emph{Text + Image (Graphviz)} & \texttt{['Portugal', 'South East England', 'Denmark', 'Estonia', 'North-West', 'Yorkshire and the Humber', 'North-West France', 'North West England', 'South', 'West Midlands']} \quad \textcolor{red}{$\bm{\times}$ (Hierarchical Visual Bottleneck)} \\
-- \emph{Text + Image (non-hierarchical Graphviz)} & \texttt{['Portugal']} \quad \textcolor{teal}{$\bm{\checkmark}$ (Correct Precision via Spatial Separation)} \\
-- \emph{Text + Image (NetworkX)} & \texttt{['Portugal']} \quad \textcolor{teal}{$\bm{\checkmark}$ (Correct Precision via Cluster Detachment)} \\
-- \emph{Text + Path Images} & \texttt{['Portugal']} \quad \textcolor{teal}{$\bm{\checkmark}$ (100\% Precision via Isolated Path Intersection)} \\
\bottomrule
\end{tabular}
\caption{\textbf{Matched Analysis for Case 3 (WebQTrn-2496).} Output progression for \texttt{Qwen3.5-35B-A3B} demonstrating how alternative spatial layouts (\texttt{NetworkX}, \texttt{non-hierarchical Graphviz}) and reasoning-path image decomposition resolve the visual bottleneck present in standard hierarchical \texttt{Graphviz} renderings.}
\label{tab:case3_details}
\end{table*}

\begin{figure*}[t]
    \centering

    \begin{subfigure}[t]{0.43\textwidth}
        \centering
        \includegraphics[width=\linewidth,height=0.27\textheight,keepaspectratio]
        {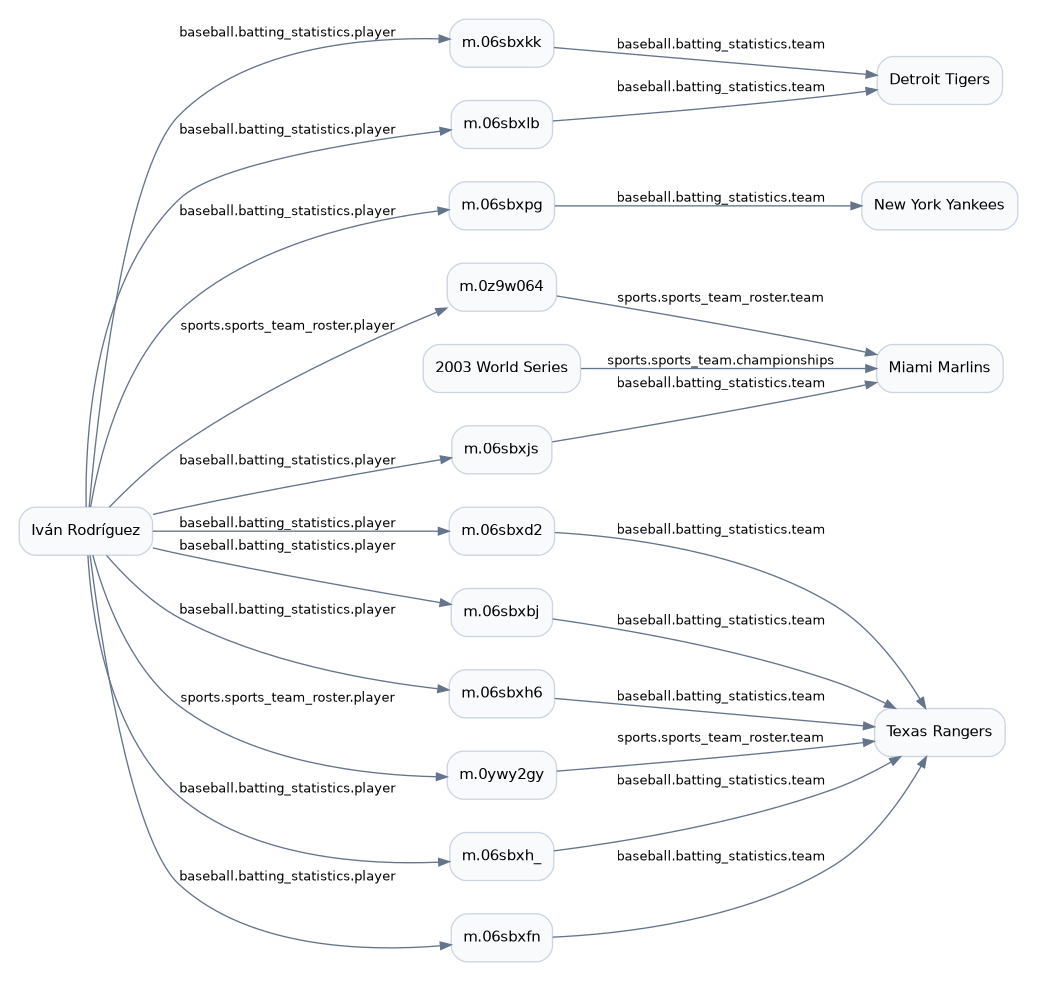}
        \caption{Graphviz}
        \label{fig:graphviz_1}
    \end{subfigure}
    \hfill
    \begin{subfigure}[t]{0.43\textwidth}
        \centering
        \includegraphics[width=\linewidth,height=0.27\textheight,keepaspectratio]
        {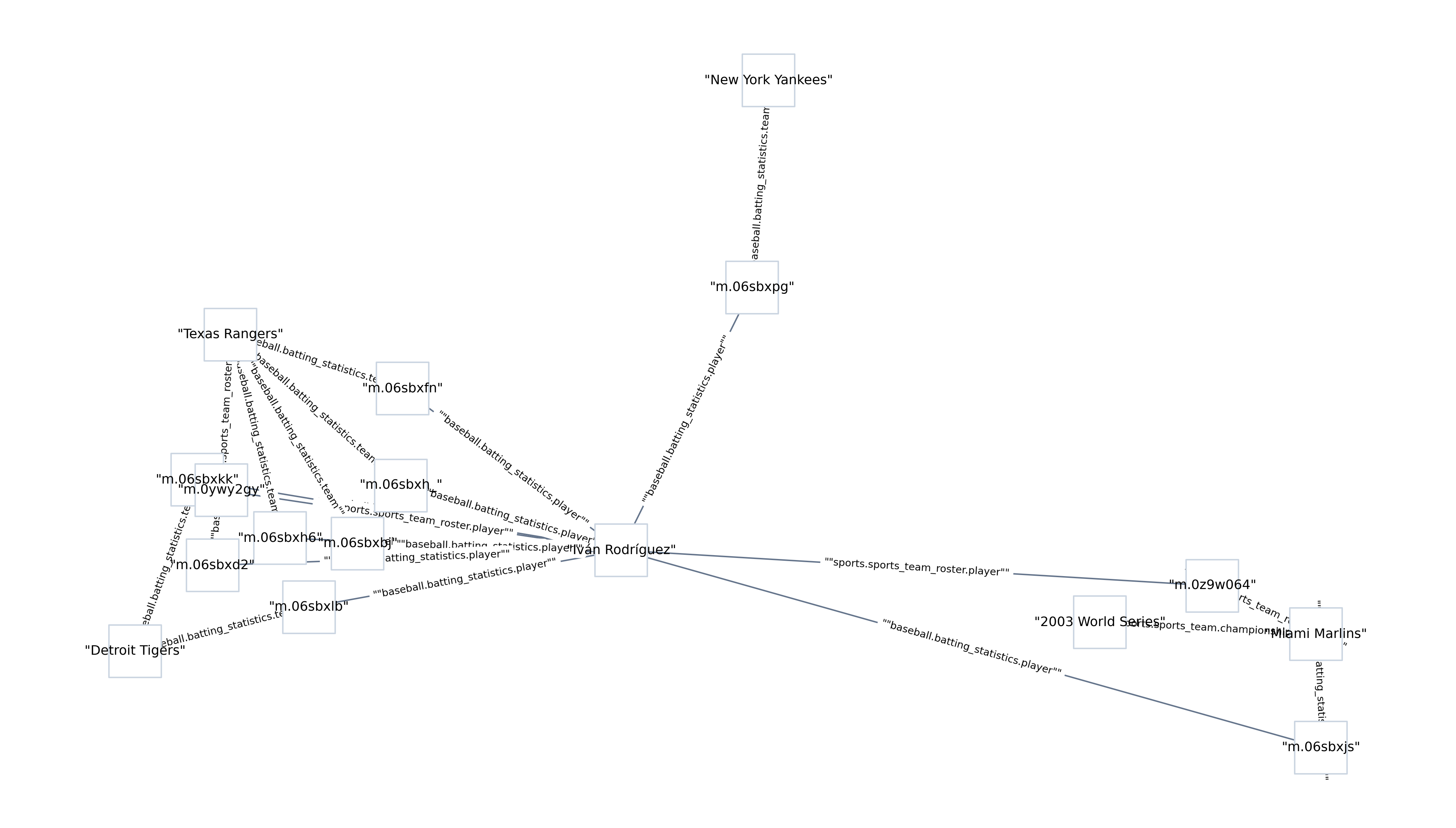}
        \caption{NetworkX}
        \label{fig:networkx}
    \end{subfigure}

    \vspace{0.5em}

    \begin{subfigure}[t]{0.43\textwidth}
        \centering
        \includegraphics[width=\linewidth,height=0.27\textheight,keepaspectratio]
        {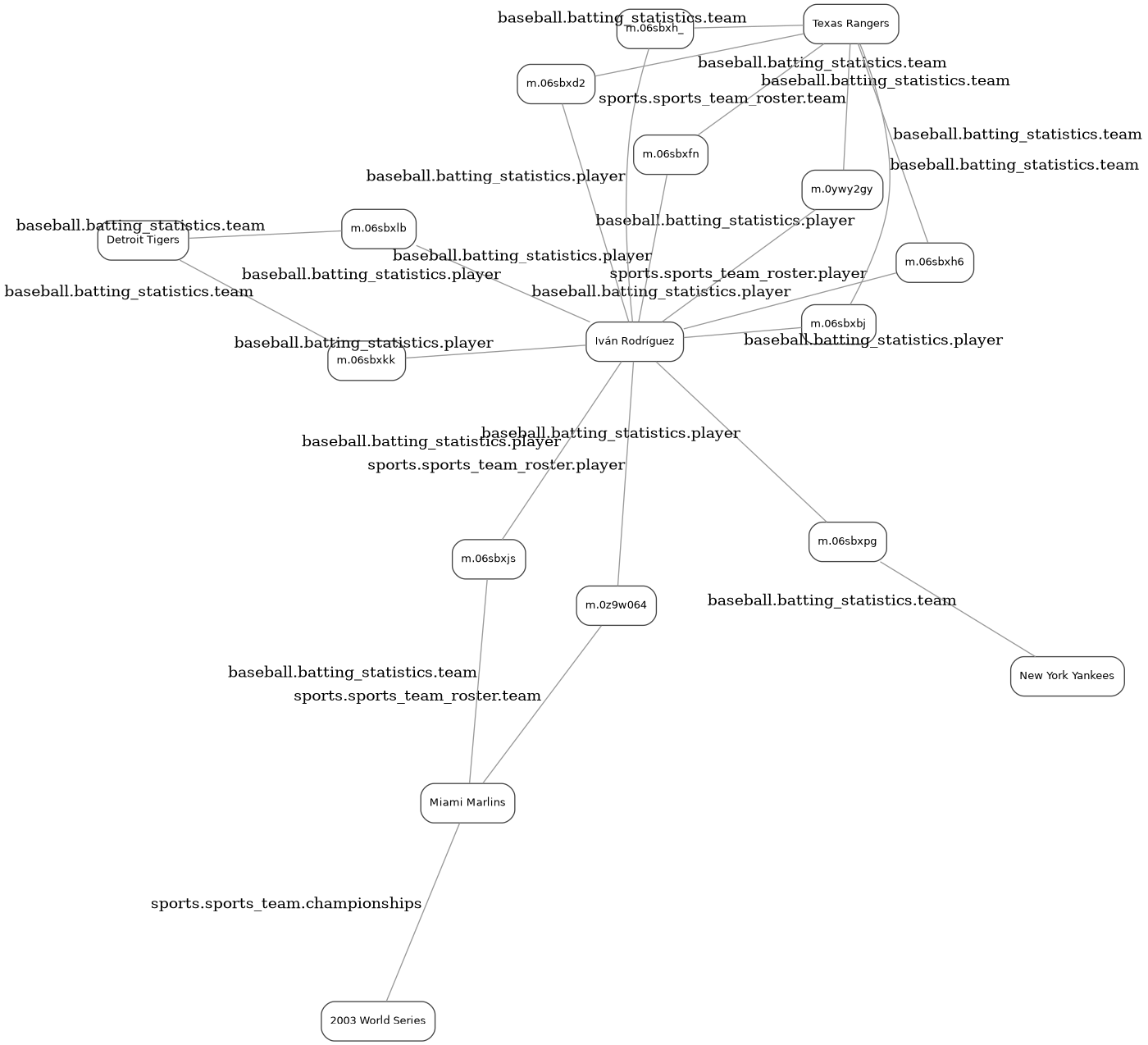}
        \caption{Graphviz with a non-hierarchical layout}
        \label{fig:graphviz-chaos}
    \end{subfigure}
    \hfill
    \begin{subfigure}[t]{0.43\textwidth}
        \centering
        \includegraphics[width=\linewidth,height=0.27\textheight,keepaspectratio]
        {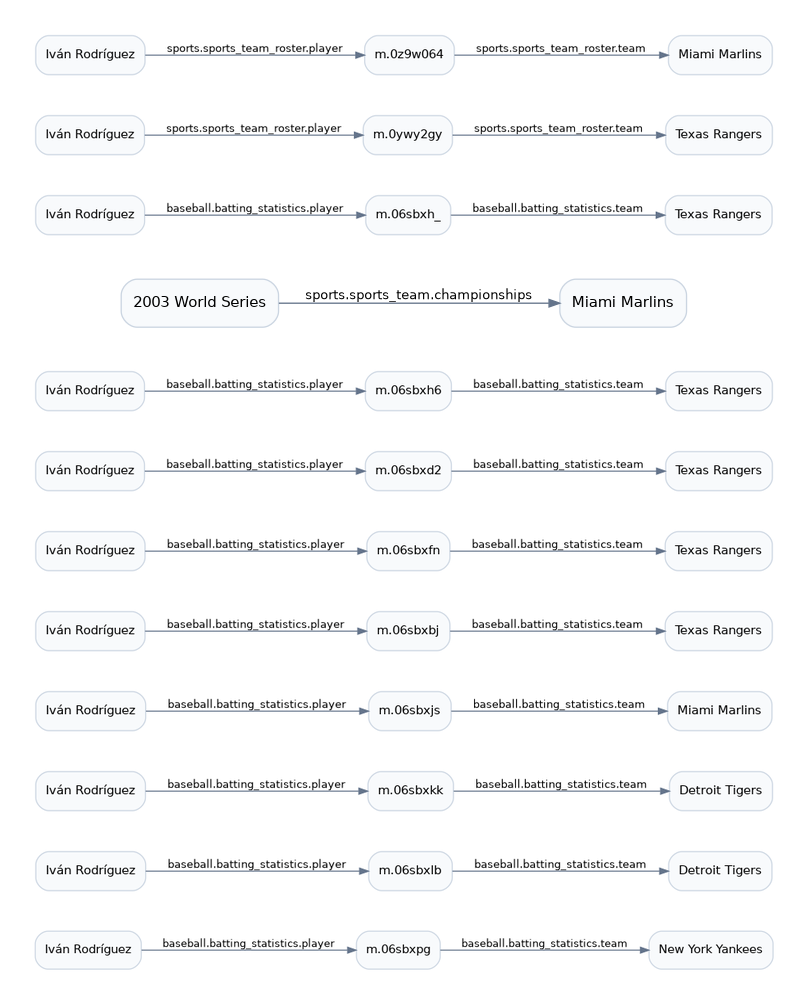}
        \caption{Reasoning paths}
        \label{fig:reasoning-paths}
    \end{subfigure}

    \caption{The four visualizations for Case 1 (\textbf{WebQTest-1319}):
    (a) Graphviz, (b) NetworkX, (c) Graphviz with a non-hierarchical layout,
    and (d) reasoning paths. }
    \label{fig:case1_graphs}
\end{figure*}

Case 3 (\autoref{tab:case3_details}) evaluates visual scalability and layout sensitivity under high graph complexity. Figure~\ref{fig:case1_graphs} presents the different visualization modes for this graph context. From this case, we can point to the following mechanisms:
\begin{itemize}
    \item \textbf{Text-Only Precision Loss:} Sequential text linearization causes the model to suffer from severe distractor overload. Failing to enforce explicit path intersection between \texttt{Portuguese Language} and \texttt{European Parliament}, the text baseline predicts 12 different regional districts alongside \emph{Portugal}.
    \item \textbf{Hierarchical Bottleneck vs. Spatial Layout Sensitivity:}
    \begin{itemize}
        \item \emph{Graphviz (Hierarchical):} Forces all 28 paths to radiate horizontally from a single root node (\texttt{European Parliament}), producing a dense visual edge stack where the overlapping labels obscure the target intersection. Consequently, the VLM reproduces the text-only error, predicting 10 districts.
        \item \emph{NetworkX \& non-hierarchical Graphviz:} By removing strict unidirectional hierarchy, force-directed positioning allows the small \texttt{Portuguese Language} sub-cluster to visually isolate itself from the \texttt{European Parliament} hub. This spatial separation enables the model to filter out non-intersecting European Parliament districts and isolate \emph{Portugal}.
    \end{itemize}
    \item \textbf{Image-Only Hallucination:} Under \emph{Image-Only} input, the model predicts \texttt{Azores} and \texttt{Madeira}. Although these are autonomous Portuguese regions, they are absent from the retrieved KB graph, demonstrating that ungrounded visual prompts risk domain hallucinations.
    \item \textbf{Path Decomposition as a Stabilizer:} Partitioning the large graph into individual path diagrams (\texttt{reasoning\_paths}) consistently achieves 100\% precision (\texttt{['Portugal']}) without relying on specific 2D force-directed spatial layouts.
\end{itemize}

\begin{figure*}[t]
    \centering

    \begin{subfigure}[t]{0.43\textwidth}
        \centering
        \includegraphics[width=\linewidth,height=0.27\textheight,keepaspectratio]
        {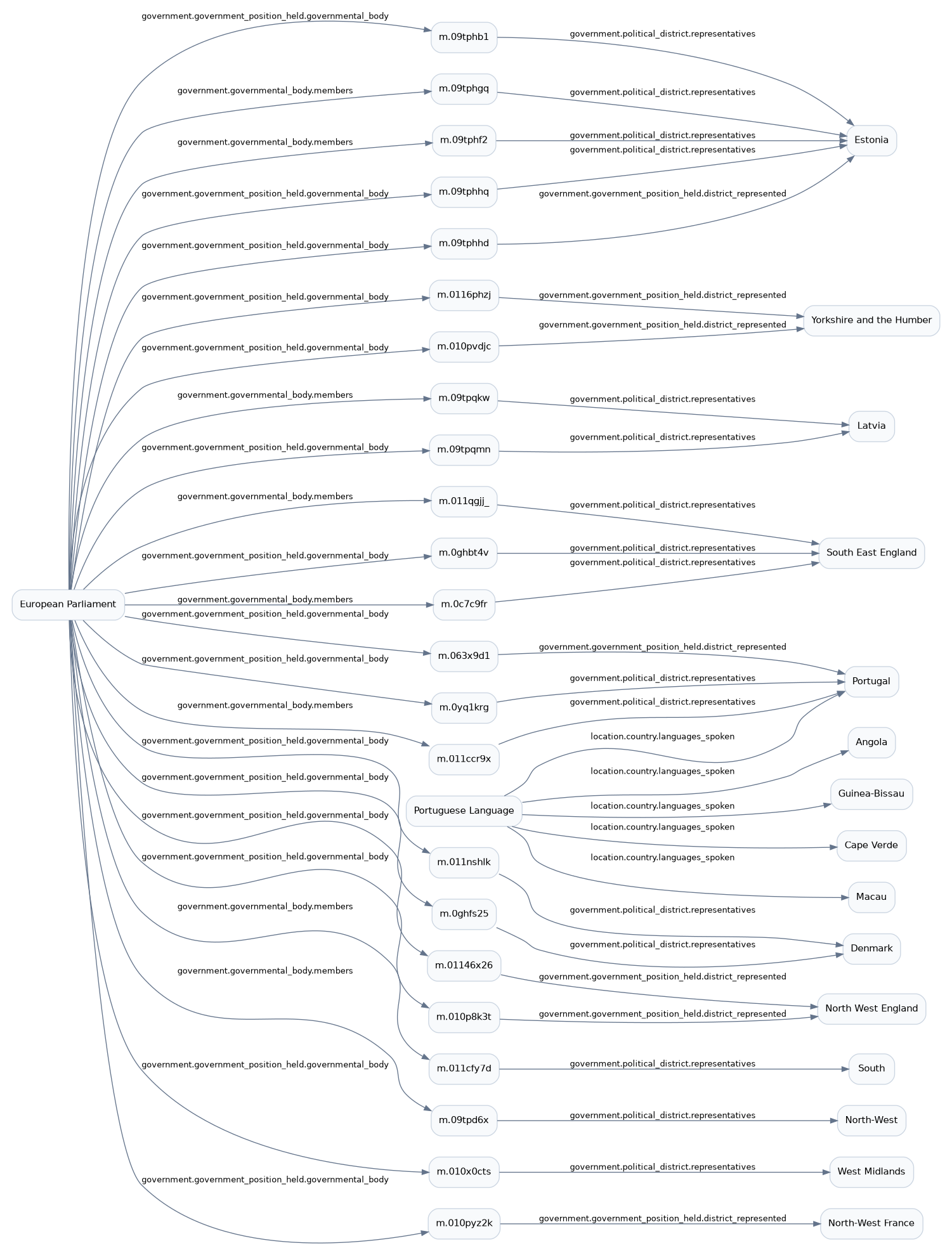}
        \caption{Graphviz}
        \label{fig:graphviz}
    \end{subfigure}
    \hfill
    \begin{subfigure}[t]{0.43\textwidth}
        \centering
        \includegraphics[width=\linewidth,height=0.27\textheight,keepaspectratio]
        {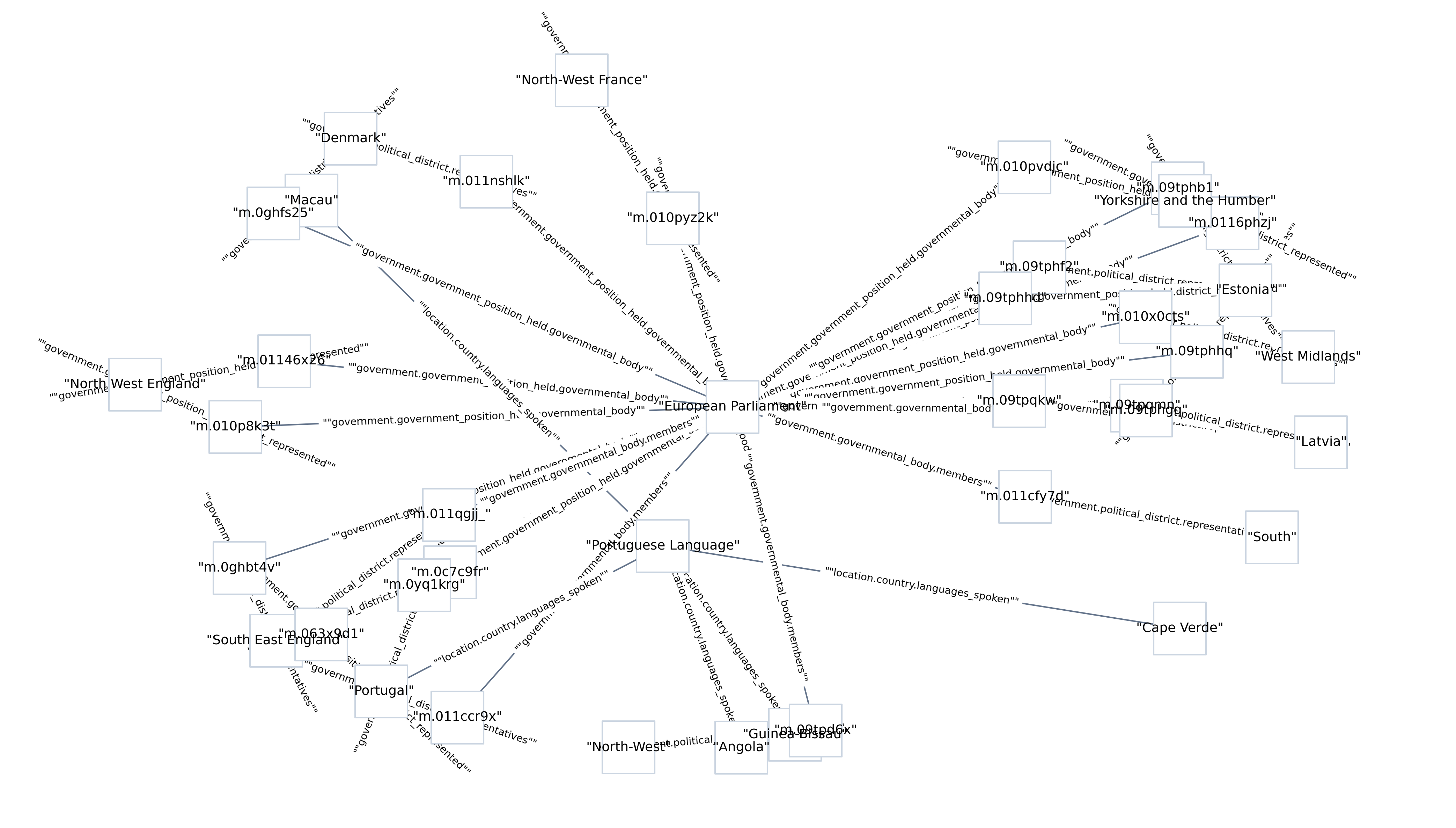}
        \caption{NetworkX}
        \label{fig:networkx}
    \end{subfigure}

    \vspace{0.5em}

    \begin{subfigure}[t]{0.43\textwidth}
        \centering
        \includegraphics[width=\linewidth,height=0.27\textheight,keepaspectratio]
        {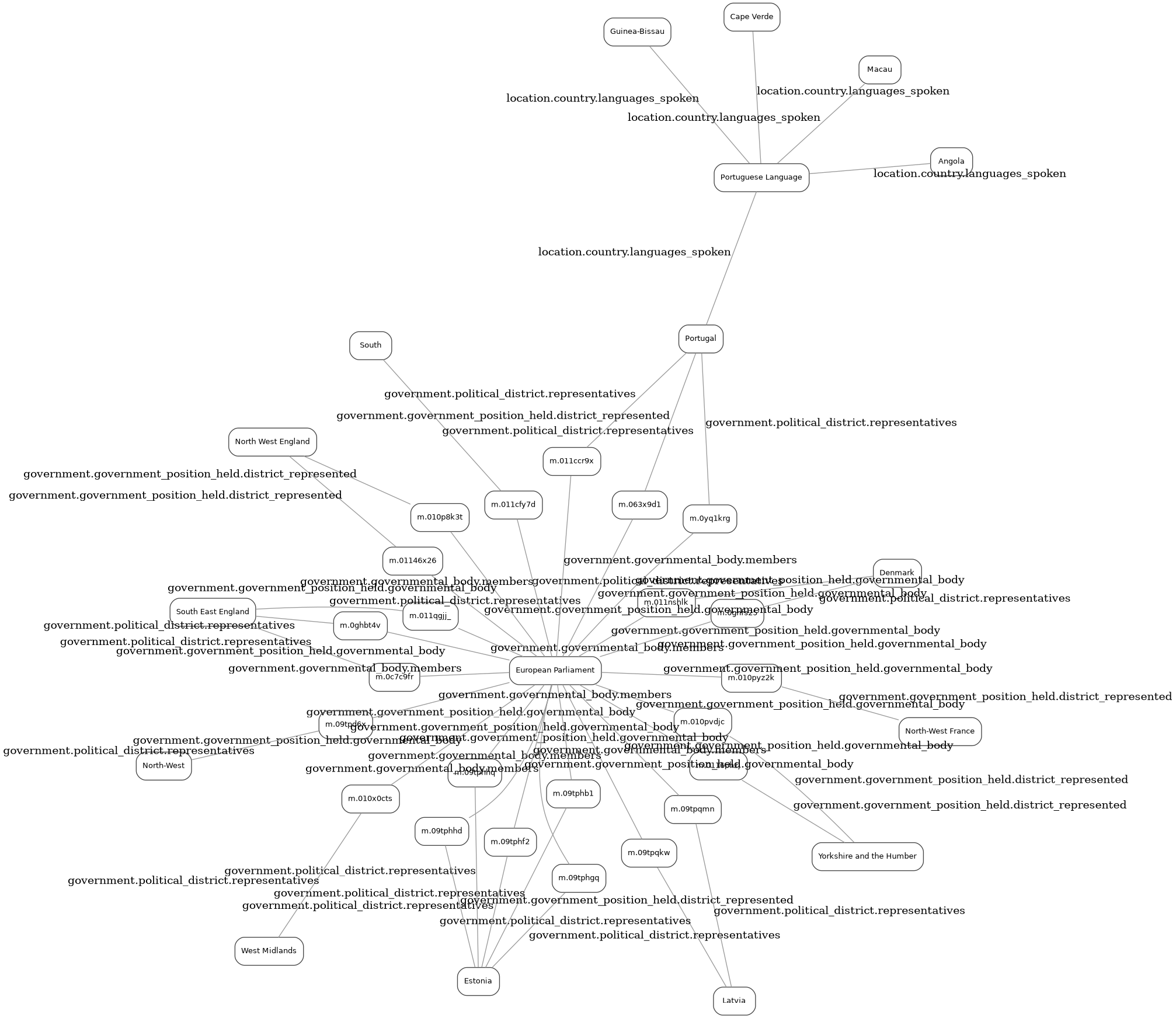}
        \caption{Graphviz with a non-hierarchical layout}
        \label{fig:graphviz-chaos}
    \end{subfigure}
    \hfill
    \begin{subfigure}[t]{0.43\textwidth}
        \centering
        \includegraphics[width=\linewidth,height=0.27\textheight,keepaspectratio]
        {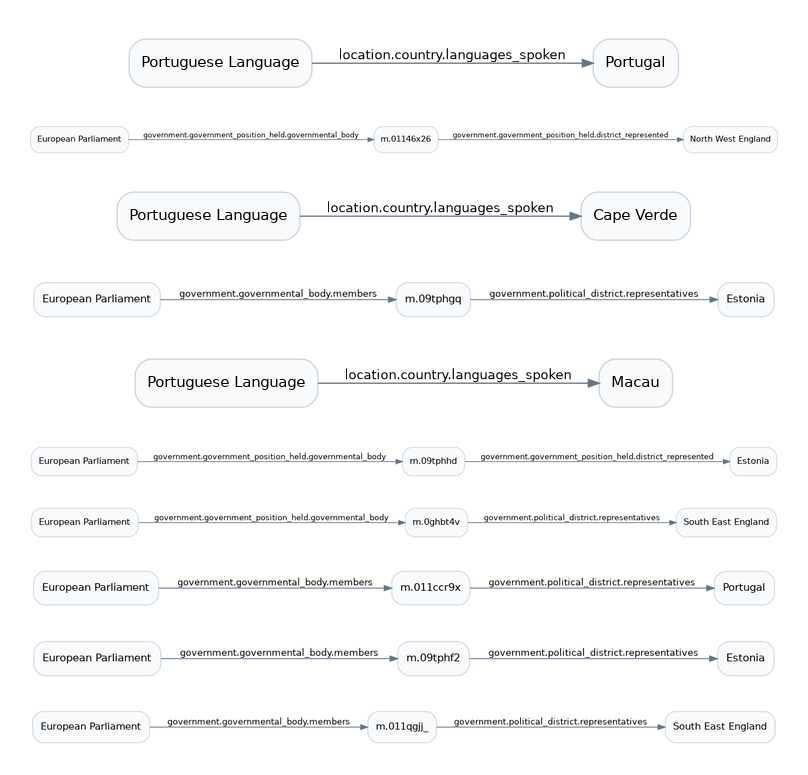}
        \caption{Excerpt of reasoning paths}
        \label{fig:reasoning-paths}
    \end{subfigure}

    \caption{The four visualizations for case 3 (\textbf{WebQTest-2496}):
    (a) Graphviz, (b) NetworkX, (c) Graphviz with a non-hierarchical layout,
    and (d) excerpt of reasoning paths. }
    \label{fig:case3_graphs}
\end{figure*}

\end{document}